# Robust tightly coupled pose estimation based on monocular vision, inertia, and wheel speed

Peng Gang, Lu Zezao, Chen Bocheng, Chen Shanliang, He Dingxin
School of Artificial Intelligence and Automation, Huazhong University of Science and Technology, Wuhan, 430074, China

*Abstract*—**The visual SLAM method is widely used for self-localization and mapping in complex environments. Visual-inertia SLAM, which combines a camera with IMU, can significantly improve the robustness and enable scale weak-visibility, whereas monocular visual SLAM is scale-invisible. For ground mobile robots, the introduction of a wheel speed sensor can solve the scale weak-visible problem and improve the robustness under abnormal conditions. In this thesis, a multi-sensor fusion SLAM algorithm using monocular vision, inertia, and wheel speed measurements is proposed. The sensor measurements are combined in a tightly coupled manner, and a nonlinear optimization method is used to maximize the posterior probability to solve the optimal state estimation. Loop detection and back-end optimization are added to help reduce or even eliminate the cumulative error of the estimated poses, thus ensuring global consistency of the trajectory and map. The wheel odometer pre-integration algorithm, which combines the chassis speed and IMU angular speed, can avoid repeated integration caused by linearization point changes during iterative optimization; state initialization based on the wheel odometer and IMU enables a quick and reliable calculation of the initial state values required by the state estimator in both stationary and moving states. Comparative experiments were carried out in room-scale scenes, building scale scenes, and visual loss scenarios. The results showed that the proposed algorithm has high accuracy, 2.2 m of cumulative error after moving 812 m (0.28%, loopback optimization disabled), strong robustness, and effective localization capability even in the event of sensor loss such as visual loss. The accuracy and robustness of the proposed method are superior to those of monocular visual inertia SLAM and traditional wheel odometers.**

***Index Terms*— Multi-sensor Fusion, Robot Pose Estimation, Simultaneous Localization and Mapping, Visual Inertia System.**

## I. INTRODUCTION

Many excellent monocular vision SLAM systems have been proposed, such as ORB-SLAM2 [1], LSD-SLAM [2], DSO [3], and SVO [4]. However, due to the drawbacks of monocular vision sensors, some limitations remain, regardless of the monocular vision SLAM algorithm used, including scale uncertainty, weak or strong light scenes, low-texture or less feature scenes, and fast motion. To this end, sensors with scale measurement capabilities and monocular vision sensors are used to perform fusion vision SLAM to increase the accuracy and robustness. Relatively stable and reliable solutions can be obtained with lasers [5]; however, this method is only suitable for large-scale scenarios, such as unmanned driving, and is unsuitable for applications with limited costs. The IMU has become a generally accepted option. However, it exhibits a non-negligible cumulative error if run for a long time [6], especially in a visually restricted condition without texture or under weak illumination, in which case the visual mile cannot be used to correct the IMU error. In [7], the scale observability of a monocular visual inertial odometer on a ground mobile robot was analyzed in detail. When the robot moves at a constant speed, due to the lack of acceleration excitation, the constraint on the scale is lost, resulting in a gradual increase in the scale uncertainty and positioning error.

An open source monocular vision inertial mileage calculation method, namely VINS-Mono, has been proposed based on tightly coupled nonlinear optimization [8,9]. By combining the IMU pre-integral measurement and visual measurement in a tightly coupled form to solve the maximum posterior probability estimation problem, we can use the nonlinear optimization method to estimate the optimal state. An open source visual inertia SLAM algorithm VINS-Fusion [10,11] was developed on the basis of VINS-Mono, supporting multiple sensor combinations (binocular camera + IMU; monocular camera + IMU; binocular camera only); this can be used for absolute pose measurement provided by GPS to further improve the accuracy of the global path.

Generally, ground mobile robots have wheel speed sensors. If the characteristics of the camera, IMU, and wheel speed sensors are fully utilized and data fusion is performed, the ability to deal with the above problems will be improved. A wheel speed inertial odometer was integrated with a monocular visual odometer based on EKF [12, 13], assuming that the robot is running on an ideal plane, and 3 DOF pose estimation was performed. The wheel speed inertial odometer uses wheel speed measurement; angular speed measurement is integrated for dead reckoning; wheel speed inertial odometer is used for EKF status prediction; and visual odometer method is used for EKF measurement update. In the above-mentioned EKF-based

"This work was supported by National Natural Science Foundation of China, No.61672244."

The next few paragraphs should contain the authors' current affiliations, including current address and e-mail. For example, F. A. Author is with the National Institute of Standards and Technology, Boulder, CO 80305 USA (e-mail: author@ boulder.nist.gov).

S. B. Author, Jr., was with Rice University, Houston, TX 77005 USA. He is now with the Department of Physics, Colorado State University, Fort Collins, CO 80523 USA (e-mail: author@lamar.colostate.edu).

T. C. Author is with the Electrical Engineering Department, University of Colorado, Boulder, CO 80309 USA, on leave from the National Research Institute for Metals, Tsukuba, Japan (e-mail: author@nrim.go.jp).

loose coupling method, when the visual odometer cannot accurately calculate the pose due to insufficient visual characteristics, some proportion of its output in the filter will decrease, resulting in ineffective visual observation. Consequently, the accuracy is reduced.

To make full use of the constraints of sensor measurement on pose estimation and improve the accuracy of pose estimation, a visual wheel speed SLAM system based on tightly coupled nonlinear optimization has been constructed [14], and the wheel speed sensor and visual odometer were integrated in a tightly coupled manner for solving the scale uncertainty of monocular vision. The optimized method was used to solve the least-squares problem corresponding to the state estimation. However, the algorithm does not consider the unreliability of wheel speed measurement. When a robot moves on uneven surfaces or in the case of wheel slip, an incorrect wheel speed measurement will seriously affect the scale accuracy and can even lead to system failure.

In [15], tightly coupled nonlinear optimization methods were used to integrate vision, inertial sensors, and wheel speed sensors and perform pose estimation. The error cost function is composed of vision errors, inertial measurement errors, and wheel speed sensor dead reckoning errors. In addition, assuming that the vehicle is moving on an approximate plane, a "soft" plane constraint term is added to the error cost function. The experiment proves that when the robot moves with constant acceleration or does not rotate, the scale of the visual inertial odometer and the direction of gravity become unobservable, and the introduction of encoder measurement and soft plane constraints significantly improves the accuracy of the visual inertial odometry of the wheeled robots.

There are few studies on multi-sensor fusion SLAM for wheeled mobile robots based on vision, inertia, and wheel speed measurements that are tightly coupled and optimized. A mature and reliable solution is required. There is no complete solution similar to ORB-SLAM. Therefore, research on tightly coupled monocular visual odometer combined with wheel speed measurement is significant.

## II. Multi-sensor State Estimation Based on Tight Coupling Optimization

The camera, IMU, and wheel odometer used in this study do not have a hard synchronization function and cannot sample the data of the IMU and wheel odometer while triggering camera sampling. To align IMU measurements with wheel odometer measurements to the camera frame, soft alignment is required. The sampling frequency of the IMU and the wheel odometer is much higher than that of the camera, so the sampling time of the image frame is used as the alignment mark. The time alignment between different sensor measurements is shown in Figure 1.

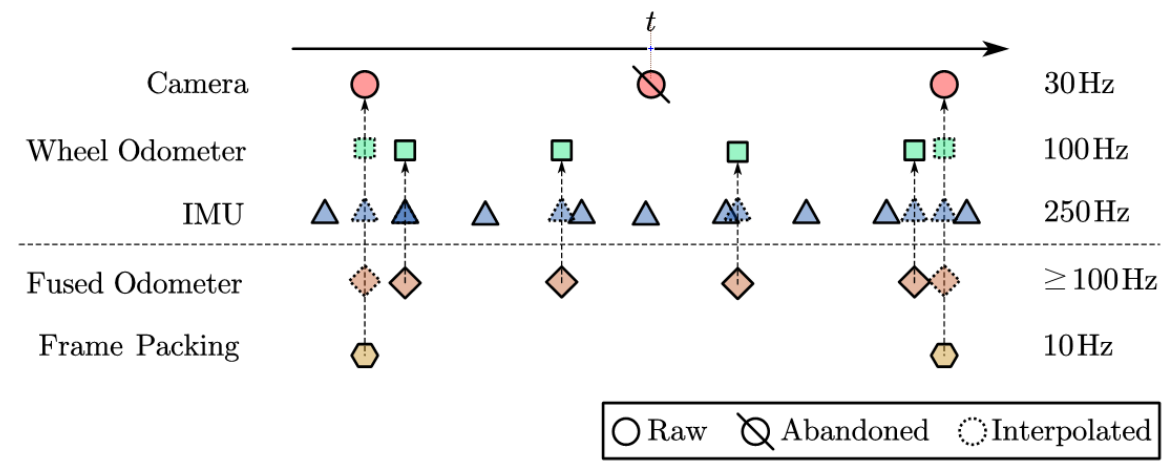

Fig. 1. Sensor pre-processing-time alignment

In Figure 1, first, the video frame obtained from the camera is down-sampled to 10 Hz, which is required by the state estimator. Subsequently, based on the original chassis wheel odometer measurement, interpolation is performed at the time position of the 10 Hz video frame. Moreover, based on the original IMU measurement, interpolation is performed at the 10 Hz video frame and the 100 Hz wheel odometer measurement time. Next, the wheel odometer measurement and its corresponding inertial measurement are packaged to generate a pre-fused wheel odometer measurement. Finally, the 10 Hz video frames, pre-fused wheel odometer measurements, and all IMU measurements are packaged into a data structure for easy state estimator processing.

In the multi-sensor state estimation process, the main data processing and analysis processes include raw sensor input, calibration compensation, data pre-processing (time alignment, pre-fused wheel odometer), pre-integration, and state estimation problem solving.

Considering that the bias of the IMU always exists, when defining the variables to be estimated $\mathcal{X}$, the IMU zero offset of each key frame is used as the variable to be estimated to participate in the optimization. The variable to be estimated $\mathcal{X}$ is defined as:

$$\mathcal{X} = \{\mathbf{x}_k, \lambda_l\}_{k\in\mathcal{K}, l\in\mathcal{L}}$$
$$\mathbf{x}_k = [\mathbf{p}^W_{B_k}, \mathbf{v}^W_{B_k}, \mathbf{q}^W_{B_k}, \mathbf{b}_{ak}, \mathbf{b}_{gk}] \tag{1}$$

In the formula, $\mathbf{x}_k$ is the IMU state in the $k$ key frame, including the position of the IMU in the world coordinate system $\mathbf{p}^W_{B_k}$, the attitude of the IMU coordinate system relative to the world coordinate system (quaternion form) $\mathbf{q}^W_{B_k}$, the IMU's position in the world coordinate system speed $\mathbf{v}^W_{B_k}$, accelerometer bias $\mathbf{b}_a$, and gyroscope bias $\mathbf{b}_g$. $\mathcal{K}$ is the key frame in the sliding window, and $\mathcal{L}$ is the feature point observed in the key frame. $\lambda_l$ is the inverse depth of the feature point in the camera coordinate system of the key frame being observed for the first time (the inverse of the Z axis coordinate).

On the basis of VINS-Mono, pre-fusion wheel odometer observations are added, so the observation $\mathcal{Z}$ used to constrain the variable $\mathcal{X}$ to be estimated is defined as:

$$\mathcal{Z} = \{\mathcal{Z}_{C_i}, \mathcal{B}_{ij}, \mathcal{O}_{ij}\}_{(i,j)\in\mathcal{K}} \tag{2}$$

Visual feature point observation $\mathcal{Z}_{C_i} = \{\hat{\mathbf{z}}_{il}\}_{l\in\mathcal{L}_i}$, containing all feature points $\mathcal{L}_i$ observed under the $i$ keyframe; IMU pre-integration observation $\mathcal{B}_{ij}=\{\hat{\mathbf{a}}_t,\hat{\boldsymbol{\omega}}_t\}_{t_i\leq t\leq t_j}$, which is obtained by integrating all IMU measurements between the $i$ key frame and the $j$ key frame; the pre-fusion wheel odometer observation $\mathcal{O}_{ij}=\{\Delta\hat{\mathbf{m}}_{\mathrm{odom}t},\hat{\mathbf{a}}_{\mathrm{avg}t},\hat{\boldsymbol{\omega}}_{\mathrm{avg}t}\}_{t_i\leq t\leq t_j}$ is obtained from all the pre-fusion wheel odometer measurement points between the $i$ key frame and the $j$ key frame.

### *A. Maximum posterior estimation*

Based on the maximum posterior estimation and Bayes' theorem, the optimal estimation problem of $\mathcal{X}$ can be transformed into:

$$\mathcal{X}^*=\underset{\mathcal{X}}{\mathrm{argmax}}\ p(\mathcal{Z}\mid\mathcal{X})p(\mathcal{X}) \tag{3}$$

Here, $p(\mathcal{Z}\mid\mathcal{X})$ is the conditional probability of observing the occurrence of $\mathcal{Z}$ under the given state $\mathcal{X}$, which can be calculated according to the observation equation and the covariance of the observation. $p(\mathcal{X})$ is the prior probability (edge probability) of the state $\mathcal{X}$, and in this paper represents the constraint on the state $\mathcal{X}$ in the sliding window by the historical observations related to the historical state that has been removed from the sliding window. Substituting the definitions of observation $\mathcal{Z}$ and state $\mathcal{X}$ into the above formula, we can get:

$$\begin{aligned}p(\mathcal{Z}\mid\mathcal{X})p(\mathcal{X})&=p(\mathcal{X})\prod_{(i,j)\in\mathcal{K}}p(\mathcal{Z}_{C_i},\mathcal{B}_{ij},\mathcal{O}_{ij}\mid\mathcal{X})\\&=p(\mathcal{X})\prod_{i\in\mathcal{K}}\prod_{l\in\mathcal{L}}p(\hat{\mathbf{z}}_{il}\mid\mathbf{x}_i,\lambda_l)\prod_{(i,j)\in\mathcal{K}}p(\{\hat{\mathbf{a}}_t,\hat{\boldsymbol{\omega}}_t\}_{t_i\leq t\leq t_j}\mid\mathbf{x}_i,\mathbf{x}_j)\\&\quad\times\prod_{(i,j)\in\mathcal{K}}p(\{\Delta\hat{\mathbf{m}}_{\mathrm{odom}t},\hat{\mathbf{a}}_{\mathrm{avg}t},\hat{\boldsymbol{\omega}}_{\mathrm{avg}t}\}_{t_i\leq t\leq t_j}\mid\mathbf{x}_i,\mathbf{x}_j)\end{aligned} \tag{4}$$

### *B. Least-squares problems*

Since finding the maximum posterior probability is equivalent to minimizing its negative logarithm, the maximum posterior estimation can be transformed into a least-squares problem. Using Mahalanobis distance to represent the degree of deviation of the residual from the covariance matrix, we can get:

$$\begin{aligned}\mathcal{X}^*=\underset{\mathcal{X}}{\mathrm{argmin}}\Big\{&\|\mathbf{r}_p-\mathbf{H}_p\mathcal{X}\|^2+\sum_{i\in\mathcal{K}}\sum_{l\in\mathcal{L}}\rho\left(\|\mathbf{r}_C(\hat{\mathbf{z}}_{il},\mathbf{x}_i,\lambda_l)\|^2_{\Sigma_{C_{il}}}\right)\\&+\sum_{(i,j)\in\mathcal{K}}\|\mathbf{r}_{\mathcal{B}}(\{\hat{\mathbf{a}}_t,\hat{\boldsymbol{\omega}}_t\}_{t_i\leq t\leq t_j},\mathbf{x}_i,\mathbf{x}_j)\|^2_{\Sigma_{\mathcal{B}_{ij}}}\\&+\sum_{(i,j)\in\mathcal{K}}\rho\left(\|\mathbf{r}_{\mathcal{O}}(\{\Delta\hat{\mathbf{m}}_{\mathrm{odom}t},\hat{\mathbf{a}}_{\mathrm{avg}t},\hat{\boldsymbol{\omega}}_{\mathrm{avg}t}\}_{t_i\leq t\leq t_j}\mid\mathbf{x}_i,\mathbf{x}_j)\|^2_{\Sigma_{\mathcal{O}_{ij}}}\right)\Big\}\end{aligned} \tag{5}$$

Here, $\|\mathbf{r}\|^2_{\Sigma}$ is the Mahalanobis distance of the residual $\mathbf{r}$ when the covariance matrix is $\Sigma$, and the Mahalanobis distance is defined as: $\|\mathbf{r}\|^2_{\boldsymbol{\Sigma}}=\mathbf{r}^T\boldsymbol{\Sigma}^{-1}\mathbf{r}$.

Because the visual measurement is easily disturbed by external factors, to improve the robustness, the Huber loss function [16] is used for the visual residual $\mathbf{r}_C$ and the wheel odometer residual $\mathbf{r}_{\mathcal{O}}$. When the Mahalanobis distance is greater than or equal to 1, or the residual error exceeds 1 standard deviation (probability of occurrence is less than approximately 32%), the gradient of the residual term for the variable $\mathcal{X}$ is 0, that is, the variable $\mathcal{X}$ is no longer constrained to avoid anomalies. The value severely affects the variables to be estimated, improving the robustness.

### *C. Visual measurement constraints*

Each visual feature point is observed again in the key frame, and a visual residual is generated. The specific process is as follows: when the visual feature point $l$ is first observed in the key frame $i$, it will be recorded and tracked. Its spatial pose is defined as a function of the key frame $i$ pose $(\mathbf{p}_{C_i},\mathbf{q}_{C_i})$ and the inverse depth $\lambda_l$ of the feature point. When the visual feature point $l$ is observed again in the key frame $j$, a visual residual term is generated. The residual term $\mathbf{r}_{C_{jl}}$ represents the error of the feature point $l$ in the position of the key frame $i$ and the position in the key frame $j$. It is also called re-projection error, which is a function of the key frame $i$ pose $(\mathbf{p}_{C_i},\mathbf{q}_{C_i})$, the key frame pose $j$, and the inverse depth $(\mathbf{p}_{C_j},\mathbf{q}_{C_j})$ of the feature point.

$$\mathbf{r}_C(\hat{\mathbf{z}}_{jl},\mathcal{X})=\mathbf{r}_{C_{jl}}(\mathbf{p}^W_{B_i},\mathbf{q}^W_{B_i},\mathbf{p}^W_{B_j},\mathbf{q}^W_{B_j},\lambda_l)=[\mathbf{b}_1\ \ \mathbf{b}_2]^T\cdot\left(\hat{\mathcal{P}}^{c_j}_l-\frac{\mathcal{P}^{c_j}_l}{\|\mathcal{P}^{c_j}_l\|}\right) \tag{6}$$

Where

$$\hat{\mathcal{P}}^{c_j}_l=\pi_c^{-1}(\hat{\mathbf{z}}_{jl})=\pi_c^{-1}\left(\begin{bmatrix}\hat{u}^{c_j}_l\\\hat{v}^{c_j}_l\end{bmatrix}\right)$$

$$\mathcal{P}^{c_j}_l=\mathbf{R}^C_B\left(\mathbf{R}^{B_j}_W\left(\mathbf{R}^W_{B_i}\left(\mathbf{R}^B_C\frac{1}{\lambda_l}\pi_c^{-1}\left(\begin{bmatrix}u^{c_u}_l\\v^{c_u}_l\end{bmatrix}\right)+\mathbf{p}^B_C\right)+\mathbf{p}^W_{B_i}-\mathbf{p}^W_{B_j}\right)-\mathbf{p}^B_C\right) \tag{7}$$

In the formula, $\hat{\mathcal{P}}^{c_j}_l$ is the position where the feature point $l$ is projected onto the unit ball in the key frame $j$, $\pi_c^{-1}$ is the back projection function, which can project the pixel coordinates into the camera coordinate system $C_j$; $\mathcal{P}^{c_j}_l$ is the position of the feature point $l$ projected on the unit ball in the key frame $i$; in order to compare the error with $\hat{\mathcal{P}}_l^{c_j}$, it needs to be transformed into the camera coordinate system $C_j$ of the key frame $j$; $[\mathbf{b}_1 \quad \mathbf{b}_2]$ are the two orthogonal base vectors on the tangent plane of the unit ball and the projection lines with the feature points in the orthogonal direction.

*D. IMU constraints*

In the visual odometer method based on bundle adjustment, the state of the carrier under each key frame is used as a variable to be optimized, and visual measurement is used to constrain. The IMU measurement between frames is added as a constraint on the optimization framework, which can improve robustness.

*1) IMU pre-integration*

To reduce the complicated operation caused by reintegration, we used the IMU pre-integration method [17] to fuse IMU measurements between two consecutive key frames. Using the Euler integral method and assuming that the derivative of each state quantity between frames is fixed, we can obtain an incremental update formula for the IMU pre-integration truth value, and the true value $\mathbf{x}$ is separated into the nominal value $\hat{\mathbf{x}}$ and the error value $\delta\mathbf{x}$. The standard Nominal update equation:

$$\hat{\mathbf{x}}_{i+1} = \begin{bmatrix} \hat{\boldsymbol{\alpha}}_{i+1}^{B_k} \\ \hat{\boldsymbol{\beta}}_{i+1}^{B_k} \\ \hat{\mathbf{q}}_{i+1}^{B_k} \\ \hat{\mathbf{b}}_{ai+1} \\ \hat{\mathbf{b}}_{gi+1} \end{bmatrix} = \begin{bmatrix} \hat{\boldsymbol{\alpha}}_i^{B_k} + \hat{\boldsymbol{\beta}}_i^{B_k}\Delta t + \frac{1}{2}\mathbf{R}\{\hat{\mathbf{q}}_i^{B_k}\}(\hat{\mathbf{a}}_{i+1} - \mathbf{b}_a)\Delta t^2 \\ \hat{\boldsymbol{\beta}}_i^{B_k} + \mathbf{R}\{\hat{\mathbf{q}}_i^{B_k}\}(\hat{\mathbf{a}}_{i+1} - \mathbf{b}_a)\Delta t \\ \hat{\mathbf{q}}_i^{B_k} \otimes \mathbf{q}\{(\hat{\boldsymbol{\omega}}_{i+1} - \mathbf{b}_g)\Delta t\} \\ \hat{\mathbf{b}}_{ai} \\ \hat{\mathbf{b}}_{gi} \end{bmatrix} \tag{8}$$

Here, the initial value: $\boldsymbol{\beta}_0 = \boldsymbol{\alpha}_0 = \mathbf{0}_{3\times 1}$. The updated equation of the error value can be written in the form of a matrix:

$$\delta\mathbf{x}_{i+1} = f(\hat{\mathbf{x}}_i, \delta\mathbf{x}_i, \hat{\mathbf{z}}_{i+1}, \boldsymbol{\eta}) = (\mathbf{I} + \mathbf{F}_i\Delta t)\delta\mathbf{x}_i + \mathbf{G}_i \begin{bmatrix} \boldsymbol{\eta}_a \\ \boldsymbol{\eta}_g \\ \boldsymbol{\eta}_{b_a} \\ \boldsymbol{\eta}_{b_g} \end{bmatrix} \Delta t$$

$$\mathbf{F}_i = \begin{bmatrix} 0 & \mathbf{I} & 0 & 0 & 0 \\ 0 & 0 & -\mathbf{R}\{\hat{\mathbf{q}}_i^{B_k}\}(\hat{\mathbf{a}}_{i+1} - \mathbf{b}_{ai+1})^\wedge & -\mathbf{R}\{\hat{\mathbf{q}}_i^{B_k}\} & 0 \\ 0 & 0 & -(\hat{\boldsymbol{\omega}}_{i+1} - \mathbf{b}_{gi+1})^\wedge & 0 & -\mathbf{I} \\ 0 & 0 & 0 & 0 & 0 \\ 0 & 0 & 0 & 0 & 0 \end{bmatrix}$$

$$\mathbf{G}_i = \begin{bmatrix} 0 & 0 & 0 & 0 \\ -\mathbf{R}\{\hat{\mathbf{q}}_i^{B_k}\} & 0 & 0 & 0 \\ 0 & -\mathbf{I} & 0 & 0 \\ 0 & 0 & \mathbf{I} & 0 \\ 0 & 0 & 0 & \mathbf{I} \end{bmatrix}$$

Based on the updated equation of the error value and the definition of covariance, the updated equation of the covariance matrix can be obtained:

$$\Sigma_{\mathcal{B}i+1} = (\mathbf{I} + \mathbf{F}_i\Delta t)\Sigma_{\mathcal{B}i}(\mathbf{I} + \mathbf{F}_i\Delta t)^T + (\mathbf{G}_t\Delta t)\mathbf{Q}(\mathbf{G}_t\Delta t)^T \tag{10}$$

Here, $\Sigma_{\mathcal{B}}$ is the covariance of the error value, $\delta\mathbf{x} \sim \mathcal{N}(\mathbf{0}, \Sigma_{\mathcal{B}})$, and the diagonal covariance matrix $\mathbf{Q} = \mathrm{diag}(\boldsymbol{\eta}_a, \boldsymbol{\eta}_g, \boldsymbol{\eta}_{b_a}, \boldsymbol{\eta}_{b_g})$ of the measurement noise. The nominal value of the IMU pre-integration term relative to the zero-biased Jacobian matrix can be calculated incrementally during the pre-integration process. The updated equation of the Jacobian matrix is $\mathbf{J}_{\mathbf{x}_0}^{\hat{\mathbf{x}}_{i+1}} = (\mathbf{I} + \mathbf{F}_i\Delta t)\mathbf{J}_{\mathbf{x}_0}^{\hat{\mathbf{x}}_i}$, and the initial value is $\mathbf{J}_0 = \mathbf{J}_{\mathbf{x}_0}^{\hat{\mathbf{x}}_0} = \mathrm{diag}(15)$.

*2) Residual term*

The IMU pre-integration processes the IMU measurement for a continuous period based on the given IMU zero offset and obtains the relative pose constraint between the initial and end states of the time period. The IMU pre-integration residual term is defined as:

$$\mathbf{r}_{\mathcal{B}}(\{\hat{\mathbf{a}}_t, \hat{\boldsymbol{\omega}}_t\}_{t_k \le t \le t_{k+1}}, \mathbf{x}_k, \mathbf{x}_{k+1}) = \begin{bmatrix} \delta\boldsymbol{\alpha}_{B_{k+1}}^{B_k} \\ \delta\boldsymbol{\beta}_{B_{k+1}}^{B_k} \\ \delta\boldsymbol{\theta}_{B_{k+1}}^{B_k} \\ \delta\mathbf{b}_{aB_{k+1}}^{B_k} \\ \delta\mathbf{b}_{gB_{k+1}}^{B_k} \end{bmatrix} = \begin{bmatrix} \mathbf{R}\{\mathbf{q}_{B_k}^W\}^T\left(\mathbf{p}_{B_{k+1}}^W - \mathbf{p}_{B_k}^W - \mathbf{v}_{B_k}^W\Delta t + \frac{1}{2}\mathrm{g}\Delta t^2\right) - \hat{\boldsymbol{\alpha}}_{B_{k+1}}^{B_k} \\ \mathbf{R}\{\mathbf{q}_{B_k}^W\}^T(\mathbf{v}_{B_{k+1}}^W - \mathbf{v}_{B_k}^W + \mathrm{g}\Delta t) - \hat{\boldsymbol{\beta}}_{B_{k+1}}^{B_k} \\ 2\{\hat{\mathbf{q}}_{B_{k+1}}^{B_k *} \otimes \mathbf{q}_{B_k}^{W*} \otimes \mathbf{q}_{B_{k+1}}^W\}_{xyz} \\ \mathbf{b}_{aB_{k+1}} - \mathbf{b}_{aB_k} \\ \mathbf{b}_{gB_{k+1}} - \mathbf{b}_{gB_k} \end{bmatrix} \tag{11}$$

The random distribution of the residual term $\mathbf{r}_{\mathcal{B}}$ conforms to $\mathcal{N}(\mathbf{0}, \Sigma_{\mathcal{B}})$, and $\Sigma_{\mathcal{B}}$ is obtained by updating the covariance equation. The IMU pre-integration provides constraints on the variables to be optimized contained in the two key frames before and after. In the process of nonlinear optimization, the essence of the constraint is to provide the direction and gradient of the variable to be optimized by calculating the Jacobian matrix of the residual of the IMU pre-integration relative to the variable to be optimized. Since the direction of gravity is obtained during the initialization of the visual inertial odometer, the gravity acceleration $\mathbf{g}^W$ is not used as a variable to be optimized.

The residual of the IMU pre-integration is compared with the Jacobian matrix $\mathbf{J}_{\mathbf{x}_k,\mathbf{x}_{k+1}}^{\mathbf{r}_{\mathcal{B}}} \in \mathbf{R}^{15\times 32}$ of the state of the two key frames before and after. The Jacobian matrix is divided into 5×10 blocks for calculation based on the variables to which each dimension belongs. Due to space limitations, we ignored some of the derivation details.

$$\partial\mathbf{r}_{\mathcal{B}} = \mathbf{J}_{\mathbf{x}_k,\mathbf{x}_{k+1}}^{\mathbf{r}_{\mathcal{B}}} \begin{bmatrix} \partial\mathbf{x}_k \\ \partial\mathbf{x}_{k+1} \end{bmatrix} \tag{9}$$

$$\begin{bmatrix} \delta\boldsymbol{\alpha}_{B_{k+1}}^{B_k} \\ \delta\boldsymbol{\beta}_{B_{k+1}}^{B_k} \\ \delta\boldsymbol{\theta}_{B_{k+1}}^{B_k} \\ \delta\mathbf{b}_{aB_{k+1}}^{B_k} \\ \delta\mathbf{b}_{gB_{k+1}}^{B_k} \end{bmatrix} = \mathbf{J}_{\mathbf{x}_k,\mathbf{x}_{k+1}}^{\mathbf{r}_B} [\delta\mathbf{p}_{B_k}^W \quad \delta\mathbf{v}_{B_k}^W \quad \delta\boldsymbol{\theta}_{B_k}^W \quad \delta\mathbf{b}_{aB_k} \quad \delta\mathbf{b}_{gB_k} \quad \delta\mathbf{p}_{B_k}^W \quad \delta\mathbf{v}_{B_k}^W \quad \delta\boldsymbol{\theta}_{B_k}^W \quad \delta\mathbf{b}_{aB_k} \quad \delta\mathbf{b}_{gB_k}]^T \tag{12}$$

*E) Wheeled odometer constraints*

On ground mobile robots, a wheel speed meter is typically used to carry out dead reckoning to obtain continuous relative poses of the robot. The continuous position and reliable scale estimation of the wheel odometer make it suitable for tasks such as path planning and navigation.

*1) Two-dimensional wheel mileage calculation method*

The two-dimensional wheel odometer has an unavoidable cumulative error, but can provide a continuous carrier trajectory. Since the wheel speed meter measures the average wheel speed during a time period, the chassis speed measurement $\hat{\mathbf{m}}_{\text{base}}$ measures the average movement speed during this time. The position and attitude update methods of the wheel speed odometer mainly include Euler points, median points, and higher-order Runge–Kutta method. Because the sampling speed of the wheel speed meter is high (1 kHz), to reduce the calculation time of the main control microcontroller, the Euler integration method is used. This is done assuming that the chassis moves in a straight line at a constant speed in the original direction during the period and rotates to a new direction at the end of the time period.

The initial state of the wheel odometer is $\hat{\mathbf{m}}_{\text{odom}0} = [p_{x0} \quad p_{y0} \quad \theta_0]^T = \mathbf{0}$. Given the previous state $\hat{\mathbf{m}}_{\text{odom}k-1} = [p_{xk-1} \quad p_{yk-1} \quad \theta_{k-1}]^T$ of the wheel odometer, the current chassis speed measurement $\hat{\mathbf{v}}_{\text{base}k} = [v_{xk} \quad v_{yk} \quad \omega_k]^T$ ,and the time difference $dt_k = t_k - t_{k-1}$, we can obtain the new wheel odometer state $\hat{\mathbf{m}}_{\text{odom}k}$ as:

$$\begin{gathered} \hat{\mathbf{m}}_{\text{odom}k} = [p_{xk} \quad p_{yk} \quad \theta_k]^T \\ \begin{bmatrix} p_{xk} \\ p_{yk} \end{bmatrix} = \begin{bmatrix} p_{xk-1} \\ p_{yk-1} \end{bmatrix} + \begin{bmatrix} \cos\theta_{k-1} & -\sin\theta_{k-1} \\ \sin\theta_{k-1} & \cos\theta_{k-1} \end{bmatrix} \begin{bmatrix} v_{xk} \\ v_{yk} \end{bmatrix} dt_k \\ \theta_k = \theta_{k-1} + \omega_k dt_k \end{gathered} \tag{13}$$

*2) Wheel odometer pre-integration.*

The wheeled mileage calculation method assumes that the robot moves on an ideal plane; however, the ground may have slopes and undulations in an actual scene. The two-dimensional wheeled mileage calculation method cannot track the movement in a three-dimensional space. Introducing the three-dimensional angular velocity measurement provided by IMU in the wheeled mileage calculation method can not only solve the problem of three-dimensional motion tracking, but also increase the accuracy and reliability of heading measurement. In this study, the wheel speed inertial mileage calculation method is used between two key frames, and the angular speed measurement of the gyroscope and the position measurement of the wheel odometer are used to measure the relative pose between the two key frames. This is called wheeled odometer pre-integration. Specifically, the wheel odometer data and the IMU data first pass through a pre-fusion step to align the two-sensor data and package them into a pre-fused wheel odometer to measure $\hat{\mathbf{m}}_{\text{fused odom}}$. We then use only the pre-fusion wheel odometer to measure, according to the wheel odometer kinematics equation, and a continuous calculation and integration is made to obtain the relative displacement over a period. Finally, the relative displacement obtained by the integration is used as the pre-integration constraint of the wheel odometer to provide the direction and gradient of variable adjustment for the nonlinear optimization process in the robot pose estimation.

The incremental update equation of the wheel odometer is:

$$\begin{cases} \mathbf{p}_{O_{i+1}}^{O_k} = \mathbf{p}_{O_i}^{O_k} + \mathbf{R}\{\mathbf{q}_{O_i}^{O_k}\}\Delta\mathbf{p}_{O_{i+1}}^{O_i} = \mathbf{p}_{O_i}^{O_k} + \mathbf{R}\{\mathbf{q}_{O_i}^{O_k}\}(\Delta\hat{\mathbf{p}}_{O_{i+1}}^{O_i} + D_{O_{k+1}}^{O_k}\boldsymbol{\eta}_{os}) \\ \mathbf{q}_{O_{i+1}}^{O_k} = \mathbf{q}_{O_i}^{O_k} \otimes \mathbf{q}_{O_{i+1}}^{O_i} = \mathbf{q}_{O_i}^{O_k} \otimes \mathbf{q}\{\mathbf{R}_B^O(\hat{\boldsymbol{\omega}}_{avgi+1} - \mathbf{b}_g - \boldsymbol{\eta}_g)\Delta t\} \\ \mathbf{b}_{gi+1} = \mathbf{b}_{gi} + \boldsymbol{\eta}_{b_g}\Delta t \end{cases} \tag{14}$$

The $\Delta\hat{\mathbf{p}}_{O_{i+1}}^{O_i} \triangleq [\Delta\hat{p}_{xi+1} \quad \Delta\hat{p}_{yi+1} \quad 0]^T$ wheel odometer with noise measurement, the initial state value $\mathbf{p}_0 = \mathbf{0}, \mathbf{q}_0 = [1 \quad 0 \quad 0 \quad 0]^T$ . $\mathbf{x} = [\mathbf{p}^{O_k} \quad \mathbf{q}^{O_k} \quad \mathbf{b}_g]^T$ is used as the pre-credit term for the wheel odometer.

The nominal weight of the wheel odometer pre-integration item can be incrementally updated based on the pre-fusion wheel odometer measurement:

$$\begin{cases} \hat{\mathbf{p}}_{O_{i+1}}^{O_k} = \hat{\mathbf{p}}_{O_i}^{O_k} + \mathbf{R}\{\hat{\mathbf{q}}_{O_i}^{O_k}\}\Delta\hat{\mathbf{p}}_{O_{i+1}}^{O_i} \\ \hat{\mathbf{q}}_{O_{i+1}}^{O_k} = \hat{\mathbf{q}}_{O_i}^{O_k} \otimes \hat{\mathbf{q}}_{O_{i+1}}^{O_i} = \hat{\mathbf{q}}_{O_i}^{O_k} \otimes \hat{\mathbf{q}}\{\mathbf{R}_B^O(\hat{\boldsymbol{\omega}}_{avgi+1} - \mathbf{b}_g)\Delta t\} \\ \hat{\mathbf{b}}_{gi+1} = \hat{\mathbf{b}}_{gi} \end{cases} \tag{15}$$

The initial value of nominal weight: $\hat{\mathbf{p}}_0 = \mathbf{0}, \hat{\mathbf{q}}_0 = [1 \quad 0 \quad 0 \quad 0]^T$.

According to the definition of the error amount of the pre-integration term of the wheel odometer, and the definition of the true and nominal values, an incremental update formula of the error amount of the pre-integration term of the wheel odometer can be obtained:

$$\begin{cases} \delta\mathbf{p}_{O_{i+1}}^{O_k} = \delta\mathbf{p}_{O_i}^{O_k} + \mathbf{R}\{\mathbf{q}_{O_i}^{O_k}\}D_{O_{k+1}}^{O_k}\boldsymbol{\eta}_{os} \\ \delta\boldsymbol{\theta}_{O_{i+1}}^{O_k} = \delta\boldsymbol{\theta}_{O_i}^{O_k} - (\hat{\boldsymbol{\omega}}_{avgi+1} - \mathbf{b}_g)^\wedge \delta\boldsymbol{\theta}_{O_i}^{O_k}\Delta t - \delta\mathbf{b}_g\Delta t + \boldsymbol{\eta}_g\Delta t \\ \delta\mathbf{b}_{gi+1} = \delta\mathbf{b}_{gi} + \boldsymbol{\eta}_{b_g}\Delta t \end{cases} \tag{16}$$

The nominal value of the wheel odometer pre-integration term depends on the pre-fusion wheel odometer measurement and the gyroscope zero offset. As the variable to be optimized, the gyroscope's zero bias needs to be continuously adjusted in the pose estimation process to reduce the residual error. Therefore, in the optimization process, the partial derivatives $\mathbf{J}_{\mathbf{b}_g}^{\mathbf{p}}$ and $\mathbf{J}_{\mathbf{b}_g}^{\boldsymbol{\theta}}$ of the nominal value of the pre-integral term of the wheel odometer with respect to the zero offset of the gyroscope need to be used.

According to the incremental update of the error value of the pre-integration term of the wheel odometer, the Jacobian matrix of the error value between the two frames before and after can be obtained as:

$$\mathbf{J}_{\delta \mathbf{x}_i}^{\delta \mathbf{x}_{i+1}} = \frac{\delta \mathbf{x}_{i+1}}{\delta \mathbf{x}_i} = \mathbf{I} + \mathbf{F}_i \tag{17}$$

According to the definition of the nominal value of the pre-integration item of the wheel odometer, the Jacobian matrix of the error value is the nominal Jacobian matrix of $\mathbf{J}_{\hat{\mathbf{x}}_i}^{\hat{\mathbf{x}}_{i+1}} = \mathbf{J}_{\delta \mathbf{x}_i}^{\delta \mathbf{x}_{i+1}}$. Therefore, according to the chain-derivation rule $\mathbf{J}_{\hat{\mathbf{x}}_0}^{\hat{\mathbf{x}}_{i+1}} = \mathbf{J}_{\hat{\mathbf{x}}_i}^{\hat{\mathbf{x}}_{i+1}} \mathbf{J}_{\hat{\mathbf{x}}_0}^{\hat{\mathbf{x}}_i}$, the update equation of the nominal value of the pre-integration term of the wheel odometer with respect to the zero-biased Jacobian matrix is:

$$\mathbf{J}_{\hat{\mathbf{x}}_0}^{\hat{\mathbf{x}}_{i+1}} = (\mathbf{I} + \mathbf{F}_i) \mathbf{J}_{\hat{\mathbf{x}}_0}^{\hat{\mathbf{x}}_i} \tag{18}$$

The initial value of the Jacobian matrix: $\mathbf{J}_0 = \mathbf{J}_{\hat{\mathbf{x}}_0}^{\hat{\mathbf{x}}_0} = \mathrm{diag}(9)$.

*3) Residual term*

Definition: In the least-squares problem of robot pose estimation, the wheel odometer residual term $\mathbf{r}_{\mathcal{O}}$ represents the error distance between the frame-to-frame relative displacement $\hat{\mathbf{p}}_{O_{k+1}}^{O_k}$ and the key frame displacement $\mathbf{p}_{O_{k+1}}^{O_k}$ in the variable to be optimized, where $\hat{\mathbf{p}}_{O_{k+1}}^{O_k}$ is the observation, and $\mathbf{p}_{O_{k+1}}^{O_k}$ is the estimator.

$$\mathbf{r}_{\mathcal{O}}\left(\{\Delta \hat{\mathbf{m}}_{\mathrm{odom}t}, \hat{\boldsymbol{\omega}}_{\mathrm{avg}t}\}_{t_k \le t \le t_{k+1}}, \mathbf{x}_k, \mathbf{x}_{k+1}\right) = \left[\delta \mathbf{p}_{O_{k+1}}^{O_k}\right] = \mathbf{p}_{O_{k+1}}^{O_k} - \hat{\mathbf{p}}_{O_{k+1}}^{O_k} \tag{19}$$

The wheel odometer residual does not include the errors $\delta \boldsymbol{\theta}_{O_{k+1}}^{O_k}$ and $\delta \mathbf{b}_{g O_{k+1}}^{O_k}$ with respect to the rotation and gyro zero offset. This is because these terms are already defined in the residual term of the IMU pre-integration. The IMU pre-integration uses the original IMU measurement as the angular velocity input, which provides higher rotational integration accuracy than the wheel odometer pre-integration measured with a lower frequency pre-fused wheel odometer.

To use the variable $\mathbf{x}_k, \mathbf{x}_{k+1}$ to represent the wheel odometer pre-integration residual term, $\mathbf{p}_{O_{k+1}}^{O_k}$ needs to be transformed:

$$\begin{aligned} \mathbf{p}_{O_{k+1}}^{O_k} &= \mathbf{R}_W^{O_k}\left(\mathbf{p}_{O_{k+1}}^W - \mathbf{p}_{O_k}^W\right) \\ &= \mathbf{R}_O^{B-1} \mathbf{R}\{\mathbf{q}_{B_k}^{W-1}\}\left(\mathbf{p}_{B_{k+1}}^W - \mathbf{p}_{B_k}^W\right) + \mathbf{R}_O^{B-1} \mathbf{R}\{\mathbf{q}_{B_k}^{W-1}\} \mathbf{R}\{\mathbf{q}_{B_{k+1}}^W\} \mathbf{p}_O^B - \mathbf{R}_O^{B-1} \mathbf{p}_O^B \end{aligned} \tag{20}$$

We obtain the residual term expressed using only the variables to be optimized and the wheel odometer pre-integration:

$$\mathbf{r}_{\mathcal{O}} = \mathbf{R}_O^{B-1} \mathbf{R}\{\mathbf{q}_{B_k}^{W-1}\}\left(\mathbf{p}_{B_{k+1}}^W - \mathbf{p}_{B_k}^W\right) + \mathbf{R}_O^{B-1} \mathbf{R}\{\mathbf{q}_{B_k}^{W-1}\} \mathbf{R}\{\mathbf{q}_{B_{k+1}}^W\} \mathbf{p}_O^B - \mathbf{R}_O^{B-1} \mathbf{p}_O^B - \hat{\mathbf{p}}_{O_{k+1}}^{O_k} \tag{21}$$

Here, $\mathbf{R}_O^B$ and $\mathbf{p}_O^B$ are the positions of the wheel odometer coordinate system relative to the IMU coordinate system, and are known constants.

As a maximum posterior problem, the robot pose estimation is transformed into a least-squares problem by introducing a covariance matrix of the residuals to transform the residuals with dimensions into a unified probability representation. The wheeled odometer residual $\mathbf{r}_{\mathcal{O}}$ obeys the covariance matrix $\Sigma_{\mathcal{O}}$ of the wheeled odometer pre-integration, $\mathbf{r}_{\mathcal{O}} \sim \mathcal{N}(\mathbf{0}, [\Sigma_{\mathcal{O}}]_{\mathbf{p}})$. Here, $[\Sigma_{\mathcal{O}}]_{\mathbf{p}}$ represents the displacement covariance in the wheel odometer pre-integration covariance matrix $\Sigma_{\mathcal{O}}$, $[\Sigma_{\mathcal{O}}]_{\mathbf{p}} = [\Sigma_{\mathcal{O}}]_{left\ right\ 3\times 3}$.

Jacobian matrix: According to the definition of the wheel odometer residual $\mathbf{r}_{\mathcal{O}}$, in the optimization process, the residual value $\mathbf{r}_{\mathcal{O}}$ will change with the adjustment of the previous key frame poses $\mathbf{p}_{B_k}^W$ and $\mathbf{q}_{B_k}^W$, and the poses $\mathbf{p}_{B_{k+1}}^W$ and $\mathbf{q}_{B_{k+1}}^W$ of the next key frame, and the gyroscope zero offset $\mathbf{b}_{gi}$ of the previous frame. To provide the necessary gradient direction for optimization, the system needs to be linearized in the current state $\mathbf{x}_k, \mathbf{x}_{k+1}$, and the ratio between the increment of the residual $\partial \mathbf{r}_{\mathcal{O}}$ and the increment of the variable to be optimized is calculated. Thus, the Jacobian matrix $\mathbf{J}_{\mathbf{x}_k, \mathbf{x}_{k+1}}^{\mathbf{r}_{\mathcal{O}}}$ is defined:

$$\begin{gathered} \partial \mathbf{r}_{\mathcal{O}} = \mathbf{J}_{\mathbf{x}_k, \mathbf{x}_{k+1}}^{\mathbf{r}_{\mathcal{O}}} \begin{bmatrix} \partial \mathbf{x}_k \\ \partial \mathbf{x}_{k+1} \end{bmatrix} \\ \left[\delta \mathbf{p}_{O_{k+1}}^{O_k}\right] = \mathbf{J}_{\mathbf{x}_k, \mathbf{x}_{k+1}}^{\mathbf{r}_{\mathcal{O}}} \left[\delta \mathbf{p}_{B_k}^W \ \ \delta \mathbf{v}_{B_k}^W \ \ \delta \boldsymbol{\theta}_{B_k}^W \ \ \delta \mathbf{b}_{aB_k} \ \ \delta \mathbf{b}_{gB_k} \ \ \delta \mathbf{p}_{B_k}^W \ \ \delta \mathbf{v}_{B_k}^W \ \ \delta \boldsymbol{\theta}_{B_k}^W \ \ \delta \mathbf{b}_{aB_k} \ \ \delta \mathbf{b}_{gB_k}\right]^T \end{gathered} \tag{22}$$

Here, $\mathbf{J}_{\mathbf{x}_k, \mathbf{x}_{k+1}}^{\mathbf{r}_{\mathcal{O}}} \in \mathbf{R}^{3\times 30}$. Because the increment is small, using the quaternion definition will produce additional degrees of freedom. The increment for the rotation state in the formula is defined as the shaft angle representation.

Because the wheel odometer residual is only related to some variables in the previous key frame state $\mathbf{x}_k$ and next key frame state $\mathbf{x}_{k+1}$, the value of the Jacobian matrix $\mathbf{J}_{\mathbf{x}_k, \mathbf{x}_{k+1}}^{\mathbf{r}_{\mathcal{O}}}$ is:

$$\mathbf{J}_{\mathbf{x}_k, \mathbf{x}_{k+1}}^{\mathbf{r}_{\mathcal{O}}} = \left[ \frac{\partial \delta \mathbf{p}_{O_{k+1}}^{O_k}}{\partial \delta \mathbf{p}_{B_k}^W} \ \ \mathbf{0} \ \ \frac{\partial \delta \mathbf{p}_{O_{k+1}}^{O_k}}{\partial \delta \boldsymbol{\theta}_{B_k}^W} \ \ \mathbf{0} \ \ \frac{\partial \delta \mathbf{p}_{O_{k+1}}^{O_k}}{\partial \delta \mathbf{b}_{gi}} \ \ \frac{\partial \delta \mathbf{p}_{O_{k+1}}^{O_k}}{\partial \delta \mathbf{p}_{B_{k+1}}^W} \ \ \mathbf{0} \ \ \frac{\partial \delta \mathbf{p}_{O_{k+1}}^{O_k}}{\partial \delta \boldsymbol{\theta}_{B_{k+1}}^W} \ \ \mathbf{0} \ \ \mathbf{0} \right]$$

$$\frac{\partial \delta \mathbf{p}_{O_{k+1}}^{O_k}}{\partial \delta \mathbf{p}_{B_{k+1}}^W} = \mathbf{R}_O^{B-1} \mathbf{R}\{\mathbf{q}_{B_k}^{W-1}\}$$

$$\frac{\partial \delta \mathbf{p}_{O_{k+1}}^{O_k}}{\partial \delta \mathbf{p}_{B_k}^W} = -\mathbf{R}_O^{B-1} \mathbf{R}\{\mathbf{q}_{B_k}^{W-1}\}$$

$$\frac{\partial \delta \mathbf{p}_{O_{k+1}}^{O_k}}{\partial \delta \boldsymbol{\theta}_{B_k}^W} = -\mathbf{R}_O^{B-1} \left[\mathbf{R}\{\mathbf{q}_{B_k}^{W-1}\}\left(\mathbf{p}_{B_{k+1}}^W - \mathbf{p}_{B_k}^W + \mathbf{R}\{\mathbf{q}_{B_{k+1}}^W\} \mathbf{p}_O^B\right)\right]^\wedge$$

$$\frac{\partial \delta \mathbf{p}_{O_{k+1}}^{O_k}}{\partial \delta \boldsymbol{\theta}_{B_{k+1}}^W} = -\mathbf{R}_O^{B-1} \mathbf{R}\{\mathbf{q}_{B_k}^{W-1}\} \mathbf{R}\{\mathbf{q}_{B_{k+1}}^W\} \left(\mathbf{p}_O^B\right)^\wedge$$

$$\frac{\partial \delta \mathbf{p}_{O_{k+1}}^{O_k}}{\partial \delta \mathbf{b}_{gi}} = -\mathbf{J}_{\mathbf{b}_g}^{\mathbf{p}}$$

(23)

*F. Marginalization and prior constraints*

In the state estimation based on the sliding window of the key frame, the state of the key frame and its related observations are constantly removed from the optimization equation. If all observations related to the removed key frames are directly discarded, the constraints of state estimation will be reduced, and the loss of valid information will lead to a decrease in the accuracy. Here, a marginalization algorithm is used, while removing the key frames, retaining the removed observations to constrain the optimization variables. According to [6], the use of the Gauss–Newton method to solve a nonlinear least-squares problem can be understood as adding an increment to the variable to be optimized; the objective function is the smallest. If the residual function $\mathbf{r}(\mathbf{x})$ is linearized at $\mathbf{x}$, and the Jacobian matrix $\mathbf{J}_{\mathbf{x}}^{\mathbf{r}}$ of the residual relative to the variable to be optimized is obtained, the nonlinear least-squares problem becomes a linear least-squares problem:

$$\min_{\delta \mathbf{x}} \|\mathbf{r}(\mathbf{x}+\delta\mathbf{x})\|^2 \Rightarrow \min_{\delta \mathbf{x}} \|\mathbf{r}(\mathbf{x})+\mathbf{J}_{\mathbf{x}}^{\mathbf{r}}\delta\mathbf{x}\|^2 \tag{24}$$

Here,

$$\|\mathbf{r}(\mathbf{x})+\mathbf{J}_{\mathbf{x}}^{\mathbf{r}}\delta\mathbf{x}\|^2 = [\mathbf{r}(\mathbf{x})+\mathbf{J}_{\mathbf{x}}^{\mathbf{r}}\delta\mathbf{x}]^T[\mathbf{r}(\mathbf{x})+\mathbf{J}_{\mathbf{x}}^{\mathbf{r}}\delta\mathbf{x}]$$

. Taking the derivative of this formula with respect to $\delta\mathbf{x}$ be 0, we can get:

$$\mathbf{J}_{\mathbf{x}}^{\mathbf{r}\,T}\mathbf{J}_{\mathbf{x}}^{\mathbf{r}}\delta\mathbf{x} = -\mathbf{J}_{\mathbf{x}}^{\mathbf{r}\,T}\mathbf{r}(\mathbf{x}) \tag{25}$$

Let $\mathbf{H}=\mathbf{J}_{\mathbf{x}}^{\mathbf{r}\,T}\mathbf{J}_{\mathbf{x}}^{\mathbf{r}}, \mathbf{b}=-\mathbf{J}_{\mathbf{x}}^{\mathbf{r}\,T}\mathbf{r}(\mathbf{x})$ ; thus, we get the incremental equation $\mathbf{H}\delta\mathbf{x}=\mathbf{b}$ , where $\mathbf{H}$ is called the Hessian matrix.

Divide the variable $\mathbf{x}$ to be optimized into the part $\mathbf{x}_a$ that needs to be removed and the part $\mathbf{x}_b$, $\delta\mathbf{x}=[\delta\mathbf{x}_a \quad \delta\mathbf{x}_b]^T$, that need to be retained, then the incremental equation becomes:

$$\begin{bmatrix}\mathbf{H}_a & \mathbf{H}_b \\ \mathbf{H}_b^T & \mathbf{H}_c\end{bmatrix}\begin{bmatrix}\delta\mathbf{x}_a \\ \delta\mathbf{x}_b\end{bmatrix}=\begin{bmatrix}\mathbf{b}_a \\ \mathbf{b}_b\end{bmatrix} \tag{26}$$

The Schur method is used to eliminate the element to obtain the solution of $\delta\mathbf{x}_b$:

$$\begin{bmatrix}\mathbf{I} & \mathbf{0} \\ -\mathbf{H}_b^T\mathbf{H}_a^{-1} & \mathbf{I}\end{bmatrix}\begin{bmatrix}\mathbf{H}_a & \mathbf{H}_b \\ \mathbf{H}_b^T & \mathbf{H}_c\end{bmatrix}\begin{bmatrix}\delta\mathbf{x}_a \\ \delta\mathbf{x}_b\end{bmatrix}=\begin{bmatrix}\mathbf{I} & \mathbf{0} \\ -\mathbf{H}_b^T\mathbf{H}_a^{-1} & \mathbf{I}\end{bmatrix}\begin{bmatrix}\mathbf{b}_a \\ \mathbf{b}_b\end{bmatrix}$$
$$\begin{bmatrix}\mathbf{H}_a & \mathbf{H}_b \\ \mathbf{0} & \mathbf{H}_c-\mathbf{H}_b^T\mathbf{H}_a^{-1}\mathbf{H}_b\end{bmatrix}\begin{bmatrix}\delta\mathbf{x}_a \\ \delta\mathbf{x}_b\end{bmatrix}=\begin{bmatrix}\mathbf{b}_a \\ \mathbf{b}_b-\mathbf{H}_b^T\mathbf{H}_a^{-1}\mathbf{b}_a\end{bmatrix} \tag{27}$$

Intercepting the second row of the above matrix, we get:

$$(\mathbf{H}_c-\mathbf{H}_b^T\mathbf{H}_a^{-1}\mathbf{H}_b)\delta\mathbf{x}_b=\mathbf{b}_b-\mathbf{H}_b^T\mathbf{H}_a^{-1}\mathbf{b}_a \tag{28}$$

In the above formula, only $\mathbf{x}_b$ is unknown, and no information in $\mathbf{H}$ and $\mathbf{b}$ is lost. This process removes the rows and columns related to $\mathbf{x}_a$ from the incremental equation, marginalizes the state $\mathbf{x}_a$ that needs to be removed, and retains the historical observation constraints on the state $\mathbf{x}_b$. When the next image frame arrives, the prior information in the above formula will be used as a prior constraint term to construct a nonlinear least-squares problem.

*G. SLAM based on multi-sensor fusion*

The multi-sensor fusion state estimator in this study uses monocular vision, IMU, and wheel odometer measurements based on feature point optical flow tracking. None of these sensors can measure the absolute pose. Therefore, the multi-sensor fusion state estimator, as a mileage calculation method, has an unavoidable cumulative error. To this end, we used key frame selection, loop detection, and back-end optimization [18,19,20] algorithms on the basis of VINS-Mono, and applied them to the multi-sensor fusion state estimator to form a complete SLAM system. Figure 2 shows a system block diagram of the SLAM method.

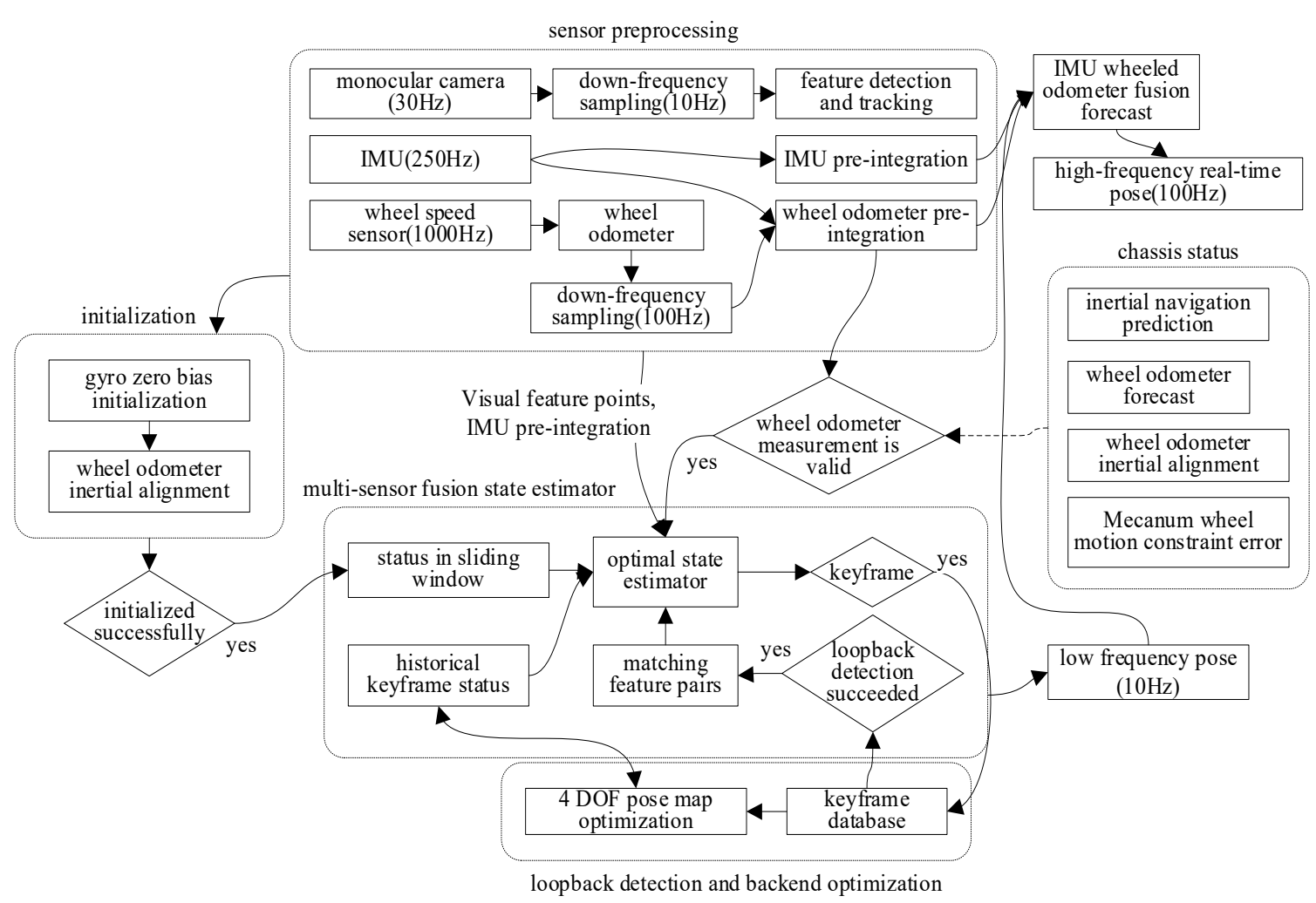

Fig. 2. Block diagram of monocular vision inertial SLAM combined with wheel speed sensor

*H. State initialization based on wheel odometer and IMU*

VINS-Mono uses multiple steps to initialize the state: gyroscope zero offset correction; initializing gravity, speed, and scale coefficients; and modifying the direction of gravity. The disadvantage of this method is that it depends on sufficient visual measurement of parallax and sufficient acceleration excitation. When there is no abnormal situation, such as skidding, the wheel odometer has better accuracy and reliability in a short distance and a short time. Compared with monocular vision, there is no scale uncertainty, and it is easier to initialize the keyframe pose, velocity, and gravity directions.

*1) Gyro zero offset initialization*

Since the gyroscope and wheel odometer measurements are on the same rigid body, the rotations of the two are the same. The relative rotation between the two key frames can be obtained through IMU pre-integration and wheel odometer pre-integration, respectively: $\mathbf{q}_{B_{k+1}}^{B_k}$ and $\mathbf{q}_{O_{k+1}}^{O_k}$. The rotation term of the pre-integration of the wheeled odometer above is also obtained through the gyro integration and has no reference value. Therefore, during the initialization process, the gyroscope pre-integration will use the heading angle of the wheel odometer for rotation integration. The rotation term $\mathbf{q}_{B_{k+1}}^{B_k}$ of the IMU pre-integration is a function of the gyroscope's bias $\mathbf{q}_{B_{k+1}}^{B_k}$. If the error between $\mathbf{q}_{B_{k+1}}^{B_k}$ and $\mathbf{q}_{O_{k+1}}^{O_k}$ is used as a constraint, the gyroscope's bias $\mathbf{b}_g$ can be estimated. Assuming that the gyro bias $\mathbf{b}_{gk}$ of each key frame during the initialization process is the same $\mathbf{b}_{gk} = \mathbf{b}_g$, the construction of the least-squares problem is as follows:

$$\mathbf{b}_g{}^* = \underset{\mathbf{b}_g}{\operatorname{argmin}} \sum_{i=1}^{k-1} \left\| \mathbf{q}_O^B \otimes \hat{\mathbf{q}}_{O_{i+1}}^{O_i}{}^{-1} \otimes \mathbf{q}_B^O \otimes \mathbf{q}_{B_{k+1}}^{B_k} \right\|^2 \tag{29}$$

Linearizing the rotation transform at $\hat{\mathbf{q}}_{B_{k+1}}^{B_k}$, we get:

$$\mathbf{q}_{B_{k+1}}^{B_k} = \hat{\mathbf{q}}_{B_{k+1}}^{B_k} \otimes \mathbf{q}\{\mathbf{J}_{\mathbf{b}_g}^{\boldsymbol{\theta}} \mathbf{b}_g\} \approx \hat{\mathbf{q}}_{B_{k+1}}^{B_k} \otimes \begin{bmatrix} 1 \\ \frac{1}{2}\mathbf{J}_{\mathbf{b}_g}^{\boldsymbol{\theta}} \mathbf{b}_g \end{bmatrix} \tag{30}$$

Here, $\mathbf{J}_{\mathbf{b}_g}^{\mathbf{q}}$ is the partial derivative of the inter-frame rotation $\mathbf{q}_{B_{k+1}}^{B_k}$ obtained by the IMU pre-integration with respect to the gyroscope zero bias $\mathbf{b}_g$. The objective function of the least-squares problem is written as:

$$\begin{gathered}\mathbf{q}_O^B \otimes \hat{\mathbf{q}}_{O_{i+1}}^{O_i}{}^{-1} \otimes \mathbf{q}_B^O \otimes \mathbf{q}_{B_{k+1}}^{B_k} = \begin{bmatrix} 1 \\ \mathbf{0} \end{bmatrix} \\ \hat{\mathbf{q}}_{B_{k+1}}^{B_k} \otimes \begin{bmatrix} 1 \\ \frac{1}{2}\mathbf{J}_{\mathbf{b}_g}^{\mathbf{q}} \mathbf{b}_g \end{bmatrix} = \mathbf{q}_O^B \otimes \hat{\mathbf{q}}_{O_{i+1}}^{O_i} \otimes \mathbf{q}_B^O \\ \begin{bmatrix} 1 \\ \frac{1}{2}\mathbf{J}_{\mathbf{b}_g}^{\mathbf{q}} \mathbf{b}_g \end{bmatrix} = \hat{\mathbf{q}}_{B_{k+1}}^{B_k}{}^{-1} \otimes \mathbf{q}_O^B \otimes \hat{\mathbf{q}}_{O_{i+1}}^{O_i} \otimes \mathbf{q}_B^O\end{gathered} \tag{31}$$

Considering only the imaginary part of the quaternion, we get:

$$\mathbf{J}_{\mathbf{b}_g}^{\mathbf{q}} \mathbf{b}_g = 2\left\{\hat{\mathbf{q}}_{B_{k+1}}^{B_k}{}^{-1} \otimes \mathbf{q}_O^B \otimes \hat{\mathbf{q}}_{O_{i+1}}^{O_i} \otimes \mathbf{q}_B^O\right\}_{xyz} \tag{32}$$

The above formula conforms to the format of $\mathbf{Hx} = \mathbf{b}$, and the Cholesky decomposition can be used to find the least-squares solution:

$$\mathbf{J}_{\mathbf{b}_g}^{\mathbf{q}}{}^T \mathbf{J}_{\mathbf{b}_g}^{\mathbf{q}} \mathbf{b}_g = 2\mathbf{J}_{\mathbf{b}_g}^{\mathbf{q}}{}^T \left\{\hat{\mathbf{q}}_{B_{k+1}}^{B_k}{}^{-1} \otimes \mathbf{q}_O^B \otimes \hat{\mathbf{q}}_{O_{i+1}}^{O_i} \otimes \mathbf{q}_B^O\right\}_{xyz} \tag{33}$$

The elasticity of the wheel, rigid connection between the wheel and the IMU, misalignment of the wheel odometer clock and the IMU clock, and calibration error of the wheel odometer rotation scale factor may lead to poor gyro work offset initialization results, if the robot rotates rapidly during the gyro work offset initialization process.

*2) Initialization of key frame speed and gravity*

Because the Mecanum wheel will tremble during the movement, and the wheeled mileage calculation method can only obtain the heading angle information, it is difficult to obtain accurate relative rotation between key frames through wheeled odometer integration. In the previous step, the zero offset of the gyroscope has been initialized, and the relative rotation between all key frames can be obtained through IMU pre-integration. Since the rotation is known, the key frame speed and gravity can be calculated by solving linear equations. Decomposing the position term $\boldsymbol{\alpha}_{B_{k+1}}^{B_k}$ and speed term $\boldsymbol{\beta}_{B_{k+1}}^{B_k}$ in the IMU pre-integration and transforming it into the form of matrix multiplication $\hat{\mathbf{z}}_{B_{k+1}}^{B_k} = \mathbf{H}_{B_{k+1}}^{B_k} \mathbf{x}_{B_{k+1}}^{B_k}$, we get:

$$\begin{bmatrix} \boldsymbol{\alpha}_{B_{k+1}}^{B_k} - \mathbf{p}_O^B + \mathbf{R}_{B_{k+1}}^{B_k} \mathbf{p}_O^B \\ \boldsymbol{\beta}_{B_{k+1}}^{B_k} \end{bmatrix} = \begin{bmatrix} -\mathbf{I}\Delta t & 0 & \frac{1}{2}\mathbf{R}_{B_0}^{B_k}\Delta t^2 & \mathbf{R}_O^B \hat{\mathbf{p}}_{O_{k+1}}^{O_k} \\ -\mathbf{I} & \mathbf{R}_{B_{k+1}}^{B_k} & \mathbf{R}_{B_0}^{B_k}\Delta t & 0 \end{bmatrix} \begin{bmatrix} \mathbf{v}_{B_k}^{B_k} \\ \mathbf{v}_{B_{k+1}}^{B_{k+1}} \\ \mathbf{g}^{B_0} \\ s \end{bmatrix} \tag{34}$$

Here, $\hat{\mathbf{z}}_{B_{k+1}}^{B_k}$ is the IMU pre-integration measurement between the key frames $B_k$ and $B_{k+1}$. The variable to be estimated related to the key frames $B_k$ and $B_{k+1}$ is defined as $\mathbf{x}_{B_{k+1}}^{B_k}$, and $\mathbf{H}_{B_{k+1}}^{B_k}$ represents the constraint between the measurement $\hat{\mathbf{z}}_{B_{k+1}}^{B_k}$ and the variable $\mathbf{x}_{B_{k+1}}^{B_k}$ to be estimated. The $s$ in the variable $\mathbf{x}_{B_{k+1}}^{B_k}$ to be estimated represents the distance of the wheel odometer with respect to the actual distance, that is, the X-axis and Y-axis scale factors of the wheel odometer $s_x, s_y$. If the IMU excitation is sufficient, it can be used to calibrate the scale factor of the wheel odometer. To ensure the reliability of initialization, $s = 1$ is defined here.

To reduce initialization errors and improve reliability, multiple key frame measurements need to be used as constraints to calculate the key frame speed and gravity direction. By combining the multiple linear equations above, we can get the least-squares problem:

$$\mathbf{x}^* = \underset{\mathbf{x}}{\operatorname{argmin}} \sum_{k \in \text{Frames}} \left\| \hat{\mathbf{z}}_{B_{k+1}}^{B_k} - \mathbf{H}_{B_{k+1}}^{B_k} \mathbf{x}_{B_{k+1}}^{B_k} \right\|^2, \mathbf{x} = \begin{bmatrix} \mathbf{v}_{B_0}^{B_0} \\ \mathbf{v}_{B_1}^{B_1} \\ \vdots \\ \mathbf{v}_{B_k}^{B_k} \\ \mathbf{g}^{B_0} \\ s \end{bmatrix}$$

(35)

Here, $\mathbf{x}$ is the variable to be estimated, and $\mathbf{x}^*$ is the optimal estimated value of $\mathbf{x}$. The program uses Cholesky decomposition to solve the least-squares problem:

$$\mathbf{H}^T\mathbf{H}\mathbf{x} = \mathbf{H}^T\hat{\mathbf{z}} \tag{36}$$

In the formula, the matrix $\mathbf{H}$ is obtained by inserting all $\mathbf{H}_{B_{k+1}}^{B_k}$ into empty columns at the corresponding positions of the unrelated variables and summing them, and $\hat{\mathbf{z}}$ is obtained by combining all $\hat{\mathbf{z}}_{B_{k+1}}^{B_k}$.

## III. EXPERIMENT

### *A. Accuracy verification experiment*

#### *1) Room-scale pose estimation experiment*

Experimental conditions: In a laboratory where objects are placed in a complex environment, as shown in Figure 3, the control robot walks through all the channels. The channel width is narrow, and the width at the narrowest point is less than 1 meter; the movement speed is maintained at approximately 0.5 m/s. Abnormal conditions during the experiment: ① Magnetic guide bars with a height of approximately 0.5 cm were fixed on the ground, and the wheels slipped slightly as they passed; ② Due to turning too close to the weakly textured wall surface, the visual tracking was completely lost several times.

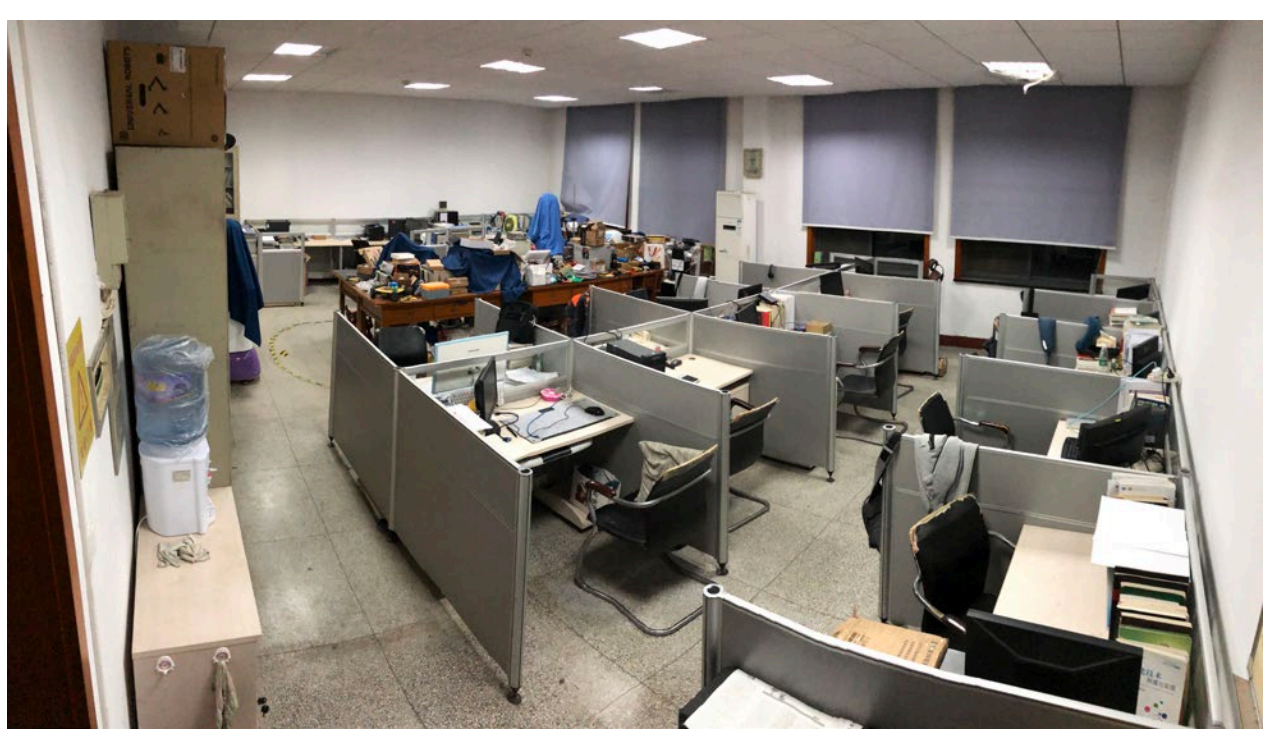

Fig. 3. Room-scale experimental environment.

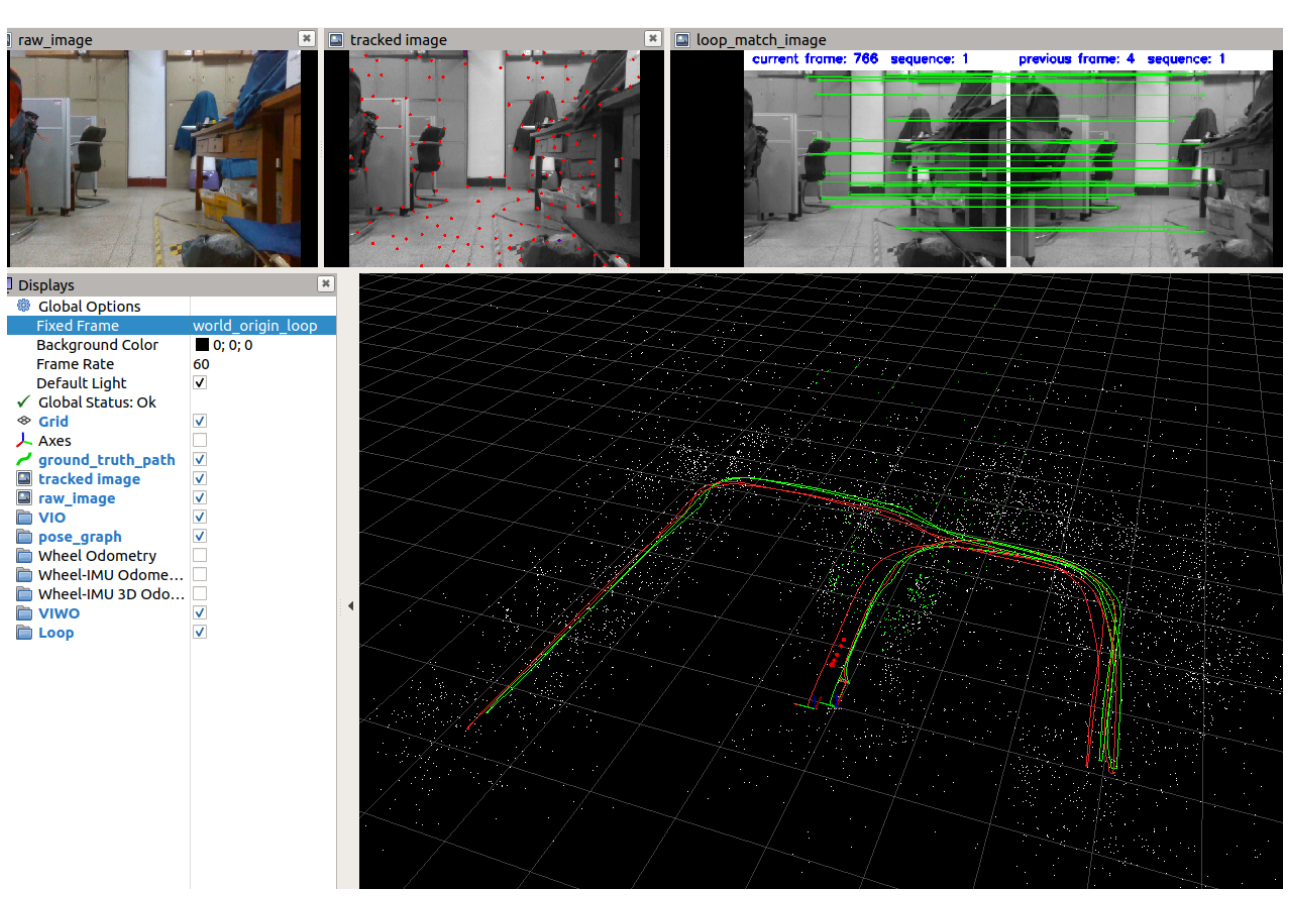

Fig. 4. RViz interface at the end of test data playback.

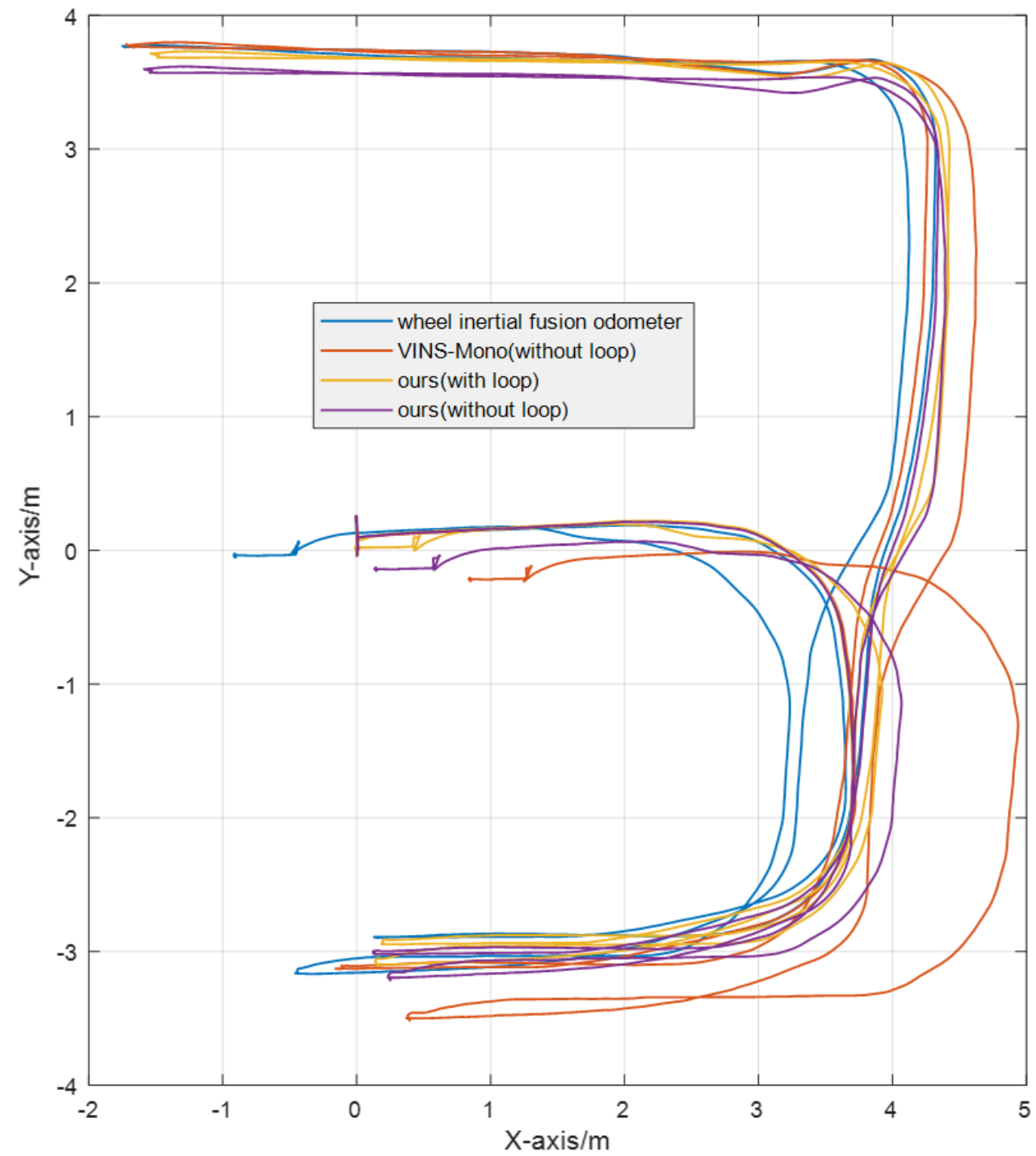

Fig. 5. Path of pose estimation at room scale (the robot starts from the origin along the positive direction of the X axis)

The path diagrams of pose estimation (Figures 5–8) start from the coordinate axis origin along the positive direction of the X axis, so the closer the path end point is to the origin of the coordinate axis, the better the position estimation effect. The first two rows in Table I provide basic information of the experiment. Because the experimental robot returns to the starting point each time and rotates to the starting direction after the end of the movement, the pose errors in the last few rows are obtained by calculating the position and angle differences between the starting and ending points of the path. The calculation method of the data in Tables II and III corresponding to the other pose estimation experiments is the same as in Table I.

TABLE I
ROOM-SCALE POSE ESTIMATION RESULTS

| chassis abnormal time | operation time | average speed | maximum speed | cumulative translation | cumulative rotation |
|---|---|---|---|---|---|
| 0.000s | 184.3s | 0.264 m/s | 0.591m /s | 51.321m | 3428.318° |
| location algorithm | X-axis error | Y-axis error | position error | position error rate | heading angle error |
| wheel odometer | -0.785 m | -0.389 m | 0.876m | 1.71% | 1.907° |
| wheel speed inertial odometer | -0.905 m | -0.040 m | 0.906m | 1.77% | −0.507° |
| VINS-Mono (without loop) | 0.851 m | -0.220 m | 0.879m | 1.71% | −0.574° |
| VINS-Mono (with loop) | 0.009 m | 0.020 m | 0.022m | 0.04% | −0.530° |
| ours (without loop) | 0.148 m | -0.143 m | 0.206m | 0.40% | −0.213° |
| ours (with loop) | -0.003 m | 0.015 m | 0.015m | 0.03% | −0.443° |

According to the data in Table I, the accuracy of posture estimation using the monocular vision inertial wheel mileage calculation method proposed in this paper is higher than that of the VINS-Mono algorithm, and the accumulated position error is only approximately 0.2 m after 51 m. We divided the cumulative position error by the cumulative translation distance to obtain the cumulative position error rate. The position error rate of the proposed algorithm is only 0.4%, which is lower than that of VINS-Mono (1%). This experiment verifies that the multi-sensor fusion mileage calculation method can perform high-precision positioning in a room-scale indoor environment, and the effect is better than that of monocular visual inertial fusion SLAM and wheel speed inertial mileage calculation method. By further combining loop detection and back-end optimization, we can achieve pose estimation with almost no bias.

*2. Pose estimation experiment on floor scale*

Experimental conditions:

① On the first floor of the building where the laboratory is located, the floor area is approximately 250×100 m;

② Control the robot: Starting from the central hall (the area of the hall is approximately 15×15 m), the robot first moves in a circle in the hall and then walks through the corridor on the west side (the width of the corridor is approximately 3–4 m). After 1 round of movement, the robot continues to walk along the corridor on the east side and finally returns to the hall and makes a circular movement to ensure that the loop detection is successful.

③ The robot's moving speed remains at approximately 1 m/s, and it does not stop when turning;

④ When passing through the corridor, the robot moves along the centerline of the corridor, and the true return path should basically coincide.

Anomalies during the experiment:

① There is a considerable amount of dust on the ground, which decreases the friction between the wheels, leading to a slight slip during rapid turns and a serious slip during left-to-right translation;

② Due to the fast movement of the robot, the picture of the rolling shutter camera continues to exhibit disturbance;

③ There are cable manhole covers in many places in the corridor, the ground surface is uneven, and there are 2–3 cm step-like undulations. The robot vibrates significantly while passing through these obstacles;

④ The corridor contains semi-open areas and closed areas, the environment brightness changes drastically, and some areas are almost completely dark;

⑤ The walls around the robot in some areas are covered with tiles and have reflections;

⑥ The corridor and hall scenes have a high degree of similarity, lacking special landmarks. Loop detection was successfully performed only when the robot passed the hall halfway and finally returned to the hall;

⑦ During the experiment, pedestrians appeared several times in the camera's field of view.

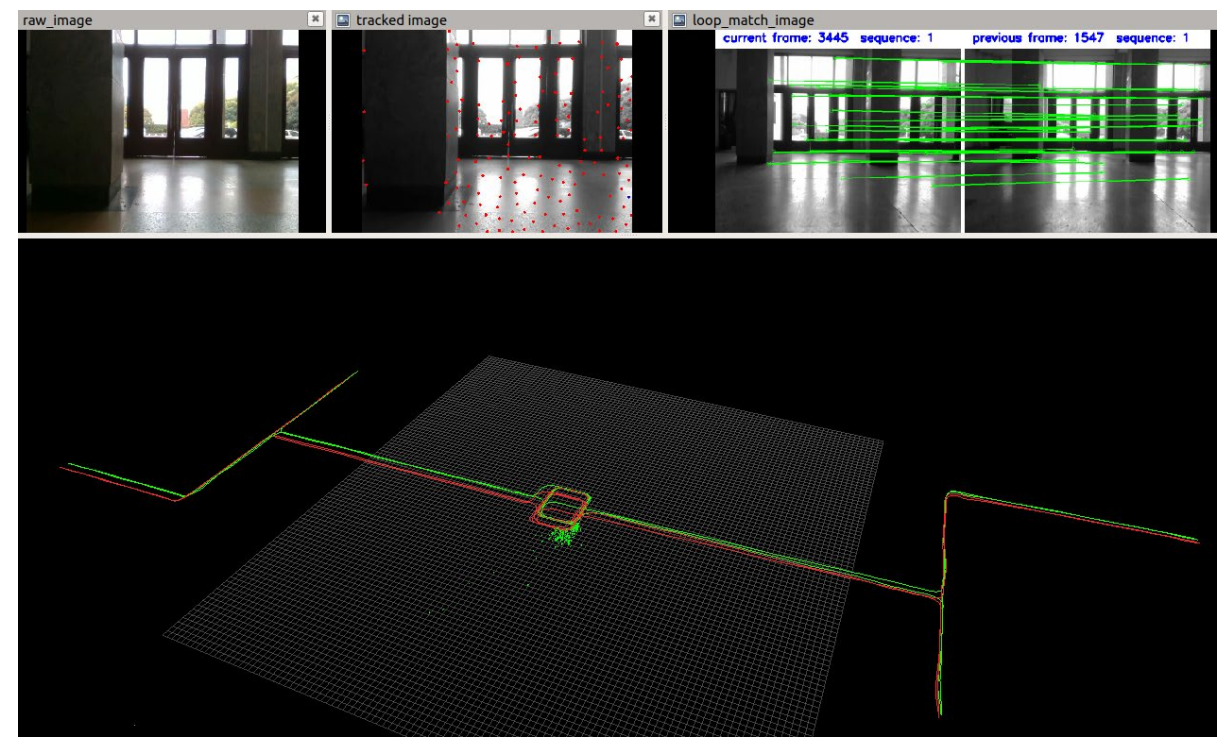

Fig. 6.  RViz interface at the end of test data playback

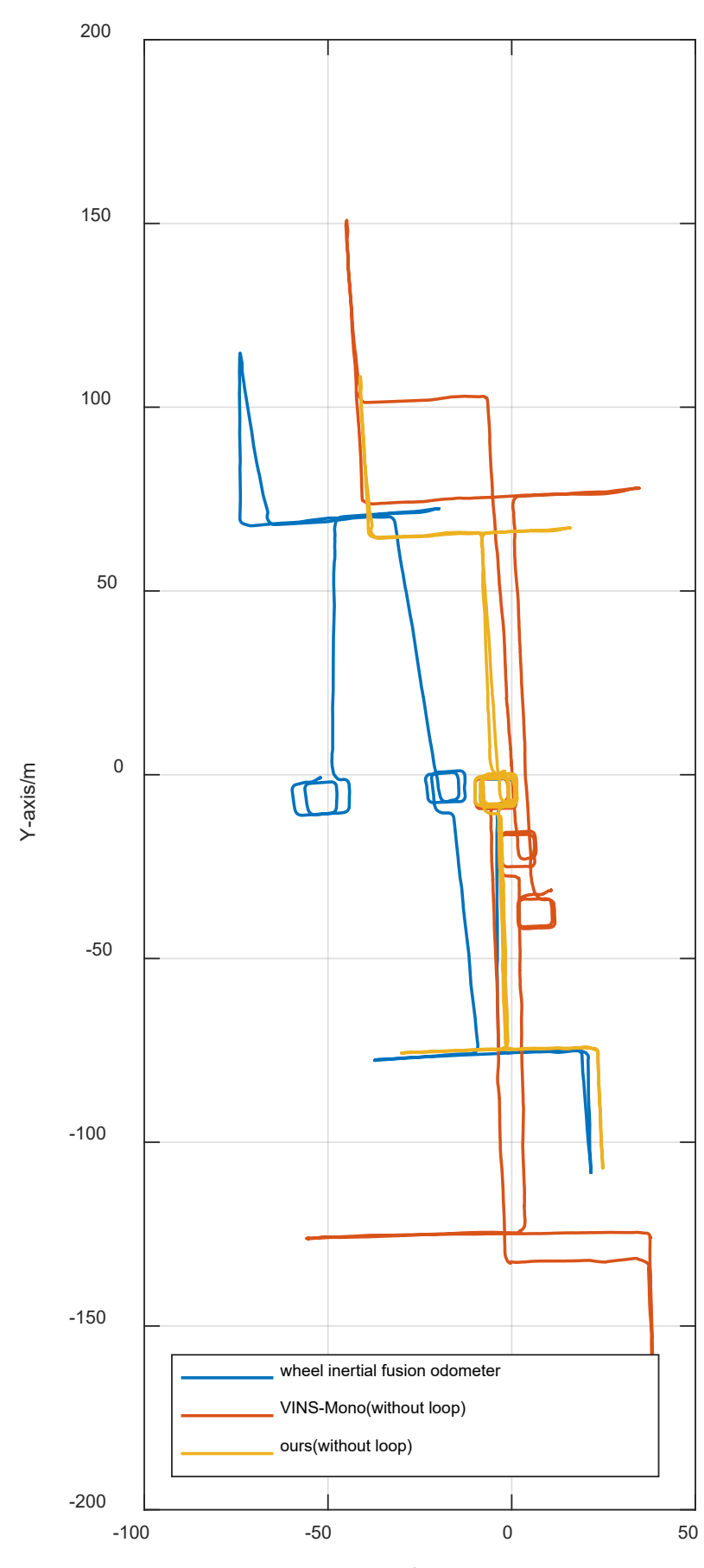

(a)

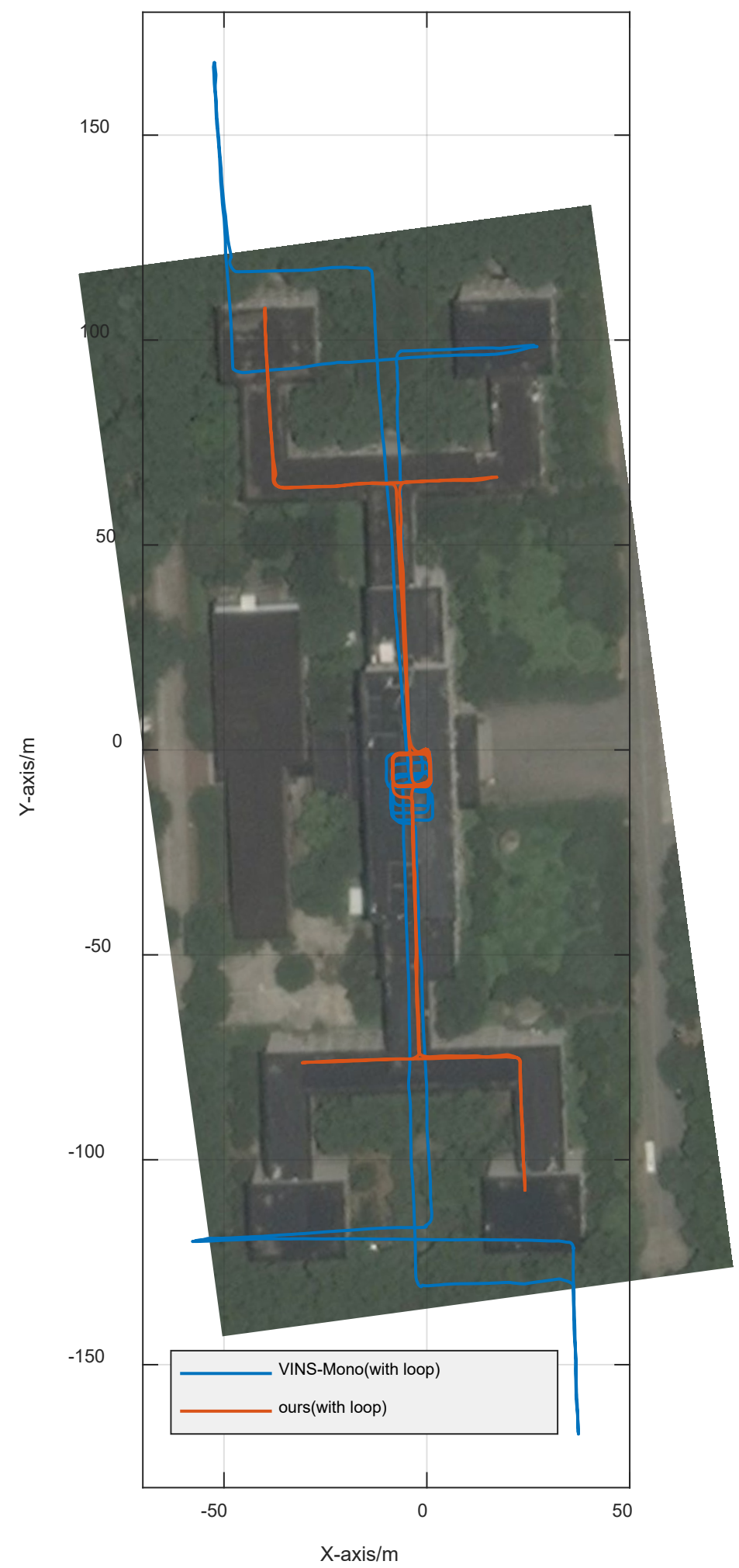

(b)

Fig. 7. Path of floor-scale pose estimation (the robot starts from the origin along the positive direction of the X axis, and (b) Projection and alignment of the satellite map of the experimental area to the path map in true scale. The satellite map is taken from Bing.com)

TABLE II
FLOOR SCALE POSE ESTIMATION RESULTS

| chassis abnormal time | operation time | average speed | maximum speed | cumulative translation | cumulative rotation |
|---|---|---|---|---|---|
| 27.171s | 896.3s | 0.885 m/s | 1.355m /s | 812.380m | 15479.221° |
| location algorithm | X-axis error | Y-axis error | position error | position error rate | heading angle error |
| wheel odometer | -83.768m | 0.295 m | 83.768 m | 10.31% | 32.868° |
| wheel speed inertial odometer | -52.108m | -0.826 m | 52.115 m | 6.42% | 6.404° |
| VINS-Mono (without loop) | 10.692m | -31.422m | 33.191 m | 4.09% | 4.080° |
| VINS-Mono (with loop) | 0.038 m | -5.281 m | 5.281m | 0.65% | 3.086° |
| ours (without loop) | -2.025 m | 0.947 m | 2.235m | 0.28% | 3.661° |
| ours (with loop) | -0.295 m | 0.165 m | 0.338m | 0.04% | 3.272° |

The cumulative positioning error of the wheel speed inertial odometer is extremely high because of the uneven ground and wheel slip. In the floor scale scenario, both the VINS-Mono algorithm with loop detection disabled and the proposed SLAM algorithm give good location results; however, there is still a cumulative positioning error that cannot be ignored. After enabling the loop detection and back-end optimization functions of the VINS-Mono algorithm and the proposed SLAM algorithm, the positioning accuracy has been significantly improved. Regardless of loop optimization, the accuracy of the proposed algorithm is better than that of VINS-Mono.

### *B. Robustness verification experiment*

To verify the robustness of the SLAM algorithm proposed in this paper, experiments are designed to test the pose estimation effect of the SLAM algorithm in the case of sensor measurement errors or even loss. Experimental conditions: In the laboratory, during the robot's movement, the visual signal is lost for approximately 15 s (the camera lens is intentionally blocked), during which the robot is kept moving and turning.

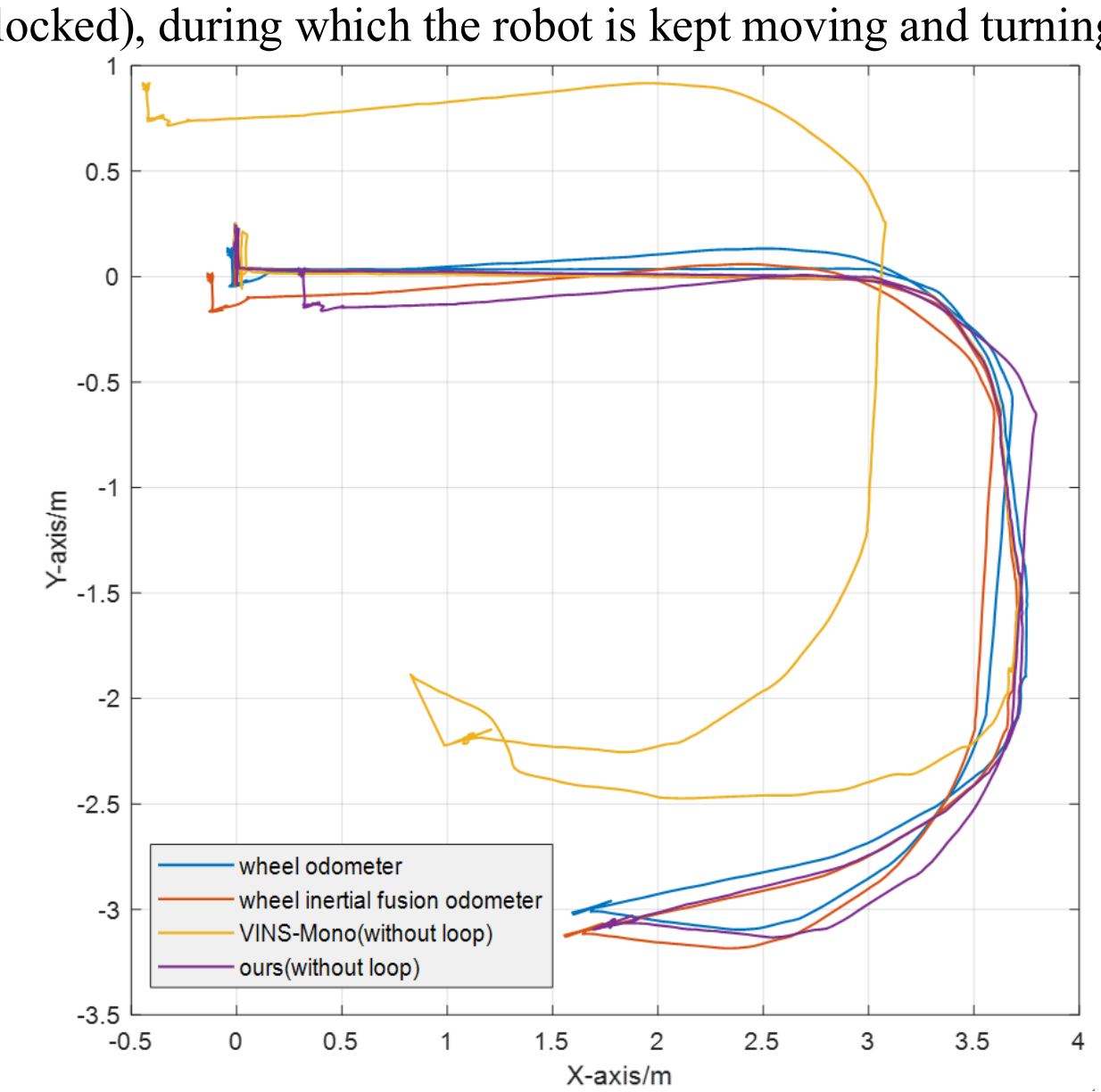

Fig. 8. Path when visual tracking is lost (the robot starts from the origin in the positive direction of the X axis).

TABLE III
POSE ESTIMATION RESULTS WHEN VISUAL TRACKING IS LOST

| chassis abnormal time | operation time | average speed | maximum speed | cumulative translation | cumulative rotation |
|---|---|---|---|---|---|
| 0.000s | 104.4s | 0.142 m/s | 0.803m /s | 17.809 s | 1110.320° |
| location algorithm | X-axis error | Y-axis error | position error | position error rate | heading angle error |
| wheel odometer | -0.042 m | 0.106 m | 0.114m | 0.64% | −1.078° |
| wheel speed | -0.137 | -0.014 | 0.137m | 0.77% | −2.811° |
| inertial odometer | m | m | | | |
| VINS-Mono (without loop) | -0.442 m | 0.890 m | 0.994m | 5.58% | −2.057° |
| VINS-Mono (with loop) | -0.020 m | 0.015 m | 0.025m | 0.14% | −1.599° |
| ours (without loop) | 0.295 m | 0.011 m | 0.295m | 1.66% | −2.883° |
| ours (with loop) | 0.030 m | -0.014 m | 0.033m | 0.19% | −2.718° |

During the experiment, the chassis did not have any abnormal conditions, the cumulative distance was short, and the accuracy of the wheel odometer was high. In the process of visual measurement loss, which the VINS-Mono relies on, the state estimator is downgraded to inertial navigation dead reckoning. The error increases rapidly, and the final positioning accuracy is poor. As shown in Figure 8, during the loss of visual features, the path of the SLAM algorithm is the same as that measured by the wheel odometer. This shows that although the positioning accuracy of the SLAM algorithm is affected by the loss of visual measurement, the pre-integration constraint of the wheel odometer can still provide absolute speed measurement; thus, the final positioning error is less than that of VINS-Mono.

Compared with the wheeled mileage calculation method, the multi-sensor fusion algorithm uses two types of sensor data. In theory, it should obtain better positioning accuracy. However, the effect in this experiment is worse than the wheeled mileage calculation method, which shows that the data fusion processing logic of the proposed algorithm needs further improvement.

## IV. Summary

For ground mobile robots, the introduction of wheel speed sensors can solve the positioning accuracy problem caused by the weak observability of monocular visual inertia SLAM and thereby improve the positioning robustness under abnormal conditions. This paper proposes a multi-sensor fusion SLAM algorithm using monocular vision, inertial measurement, and wheel speed measurement. A tightly coupled multi-sensor fusion state estimator based on the maximum posterior probability is used as the core. The visual feature points are the variables to be estimated, and the sensor is used to measure the residuals to construct a nonlinear least-squares problem, which is then solved using an optimized method. Finally, experiments were performed in room-scale scenes, floor-scale scenes, and visual loss scenarios to verify the accuracy and robustness of the algorithm.

## References

[1] R. Mur-Artal, J. M. Montiel, and J. D. Tardos, "ORB-SLAM: a versatile and accurate monocular SLAM system," *IEEE Trans. Robot*. vol. 31. no. 5, pp. 1147-1163, 2015.

[2] J. Engel, T. Schöps, and D. Cremers, "LSD-SLAM: Large-scale direct monocular SLAM," *European Conference on Computer Vision*. Zurich, Switzerland: Springer, Cham, 2014, pp. 834-849.

[3] J. Engel, V. Koltun, and D. Cremers, "Direct sparse odometry," *IEEE Trans. Pattern Anal. Mach Intelligence*, vol. 40, no. 3, pp. 611-625, 2018.

[4] C. Forster, M, Pizzoli, and D. Scaramuzza, "SVO: Fast semi-direct monocular visual odometry," *2014 IEEE International Conference on Robotics and Automation*, Hong Kong, China: IEEE, 2014: 15-22.

[5] M. Valente, C. Joly, and A. de La Fortelle, "Evidential SLAM fusing 2D laser scanner and stereo camera," Unmanned Syst. vol. 7, no. 3, 149-159, 2019.

[6] D. Unsal, and K. Demirbas, "Estimation of deterministic and stochastic IMU error parameters," *Proceedings of the 2012 IEEE/ION Position, Location and Navigation Symposium*, *IEEE*, 2012, pp. 862-868.

[7] K. J. Wu *et al*., "VINS on wheels," *2017 IEEE International Conference on Robotics and Automation*, Singapore: IEEE, 2017, pp. 5155-5162.

[8] P. Li *et al*., "Monocular visual-inertial state estimation for mobile augmented reality," 2017 *IEEE International Symposium on Mixed and Augmented Reality (ISMAR)*. Nantes, France: IEEE, 2017, pp. 11-21.

[9] T. Qin, P. Li, and S. Shen, "VINS-Mono: A robust and versatile monocular visual-inertial state estimator," *IEEE Trans. Robot*. vol. 34, no. 4, pp. 1004-1020, 2018.

[10] T. Qin *et al*., "A general optimization-based framework for global pose estimation with multiple sensors," arXiv preprint arXiv:1901.03642, 2019.

[11] T. Qin *et al*., "A general optimization-based framework for local odometry estimation with multiple sensors," arXiv preprint arXiv:1901.03638, 2019.

[12] T. B. Karamat *et al,* "Novel EKF-based vision/inertial system integration for improved navigation," *IEEE Trans. Instrumentation Meas*, vol. 67, no. 1, pp. 116-125, 2018.

[13] J. Qian *et al*., "The design and development of an omni-directional mobile robot oriented to an intelligent manufacturing system," *Sensors*, vol. 17, no. 9, p. 2073, 2017.

[14] J. Wang *et al*., "Visual SLAM incorporating wheel odometer for indoor robots," *2017 36th Chinese Control Conference* (*CCC*). Dalian, China, IEEE 2017: 5167-5172.

[15] F. Zheng, H. Tang, and Y. H. Liu, "Odometry-vision-based ground vehicle motion estimation with SE(2)-constrained SE(3) poses," *IEEE Trans. Cybernetics*, vol. 49, no. 7, pp. 2652-2663, 2018.

[16] P. J. Huber, "Robust Estimation of a Location Parameter," New York, USA: Springer, 1992.

[17] C. Forster *et al*., "IMU preintegration on manifold for efficient visual-inertial maximum-a-posteriori estimation," *2015 Robotics Science and Systems*. Roma, Italy: Georgia Institute of Technology, 2015, pp. 1-20.

[18] B. D. Lucas, and T. Kanade, "An iterative image registration technique with an application to stereo vision," *IJCAI'81 Proceedings of the 7th International Joint Conference on Artificial intelligence*. Vancouver, Canada: Morgan Kaufmann Publishers Inc, 1981, pp. 674-679.

[19] M. Calonder *et al*., "Brief: Binary robust independent elementary features," *2010 European Conference on Computer Vision*, Berlin, Germany: Springer, 2010, pp. 778-792.

[20] D. Gálvez-López, and J. D. Tardos, "Bags of binary words for fast place recognition in image sequences," *IEEE Trans. Robotics*, vol. 28, no. 5, pp. 1188-1197, 2012.

**Peng Gang** (1973-) received the Ph.D. degree in engineering from Huazhong University of Science and Technology(HUST), Wuhan, China in 2002.Currently, he is an associate professor in the Department of Automatic Control, School of Artificial Intelligence and Automation, Huazhong University of Science and Technology. A backbone teacher, a member of the Intelligent Robot Professional Committee of the Chinese Artificial Intelligence Society, a member of the China Embedded System Industry Alliance and the China Software Industry Embedded System Association, a senior member of the Chinese Electronics Association, and a member of the Embedded Expert Committee.

**Lu Zezao** (1994-) received B Eng. degree in automation from Center South University, Changsha, China, in 2016. He received master degree at the Department of Automatic Control, School of Artificial Intelligence and Automation (AIA), HUST. His research interests are intelligent robots and perception algorithms.

**Chen Bocheng** (1996-) received B Eng. degree in automation from Chongqing University, Chongqing, China, in 2018. He is currently a graduate student at the Department of Automatic Control, School of AIA, HUST. His research interests are intelligent robots and perception algorithms. **Corresponding author, Email: cbc@hust.edu.cn**.

**Chen Shanliang** (1993-) received his B Eng. degree in automation from Wuhan University of Science and Technology, Wuhan, China, in 2018. He is currently a graduate student at the Department of Automatic Control, School of AIA, HUST. His research interests are intelligent robots and perception algorithms.